\documentclass[10pt,twocolumn,letterpaper]{article}

\usepackage{iccv}
\usepackage{times}
\usepackage{epsfig}
\usepackage{graphicx}
\usepackage{amsmath}
\usepackage{amssymb}
\usepackage[accsupp]{axessibility}  

\usepackage{pifont}
\newcommand{\cmark}{\ding{51}}
\newcommand{\xmark}{\ding{55}}
\usepackage{comment}
\usepackage{multirow}
\usepackage[table, dvipsnames]{xcolor}
\usepackage[toc,page]{appendix}
\usepackage{enumitem}
\usepackage{nameref}

\usepackage{amsthm}
\theoremstyle{definition}
\newtheorem{definition}{Definition}
\newcommand{\innerproduct}[2]{\langle #1, #2 \rangle}
\newtheorem{theorem}{Theorem} 

\DeclareMathOperator*{\argmax}{arg\,max}

\usepackage[breaklinks=true,bookmarks=false]{hyperref}

\iccvfinalcopy 

\def\iccvPaperID{****} 

\ificcvfinal\fi

\begin{document}

\title{Inducing Neural Collapse to a Fixed Hierarchy-Aware Frame\\ for Reducing Mistake Severity}

\author{Tong Liang, Jim Davis\\
Ohio State University\\
Columbus, Ohio 43210\\
{\tt\small \{liang.693, davis.1719\}@osu.edu}
}

\maketitle
\ificcvfinal\thispagestyle{empty}\fi

\begin{abstract}
   There is a recently discovered and intriguing phenomenon called Neural Collapse: at the terminal phase of training a deep neural network for classification, the within-class penultimate feature means and the associated classifier vectors of all flat classes collapse to the vertices of a simplex Equiangular Tight Frame (ETF). Recent work has tried to exploit this phenomenon by fixing the related classifier weights to a pre-computed ETF to induce neural collapse and maximize the separation of the learned features when training with imbalanced data. In this work, we propose to fix the linear classifier of a deep neural network to a Hierarchy-Aware Frame (HAFrame), instead of an ETF, and use a cosine similarity-based auxiliary loss to learn hierarchy-aware penultimate features that collapse to the HAFrame. We demonstrate that our approach reduces the mistake severity of the model's predictions while maintaining its top-1 accuracy on several datasets of varying scales with hierarchies of heights ranging from 3 to 12. Code:~\href{https://github.com/ltong1130ztr/HAFrame}{\textcolor{blue}{https://github.com/ltong1130ztr/HAFrame}}.
\end{abstract}


\section{Introduction}

\begin{figure}
    \centering
    \begin{tabular}{c c}
    \includegraphics[height=1.1in]{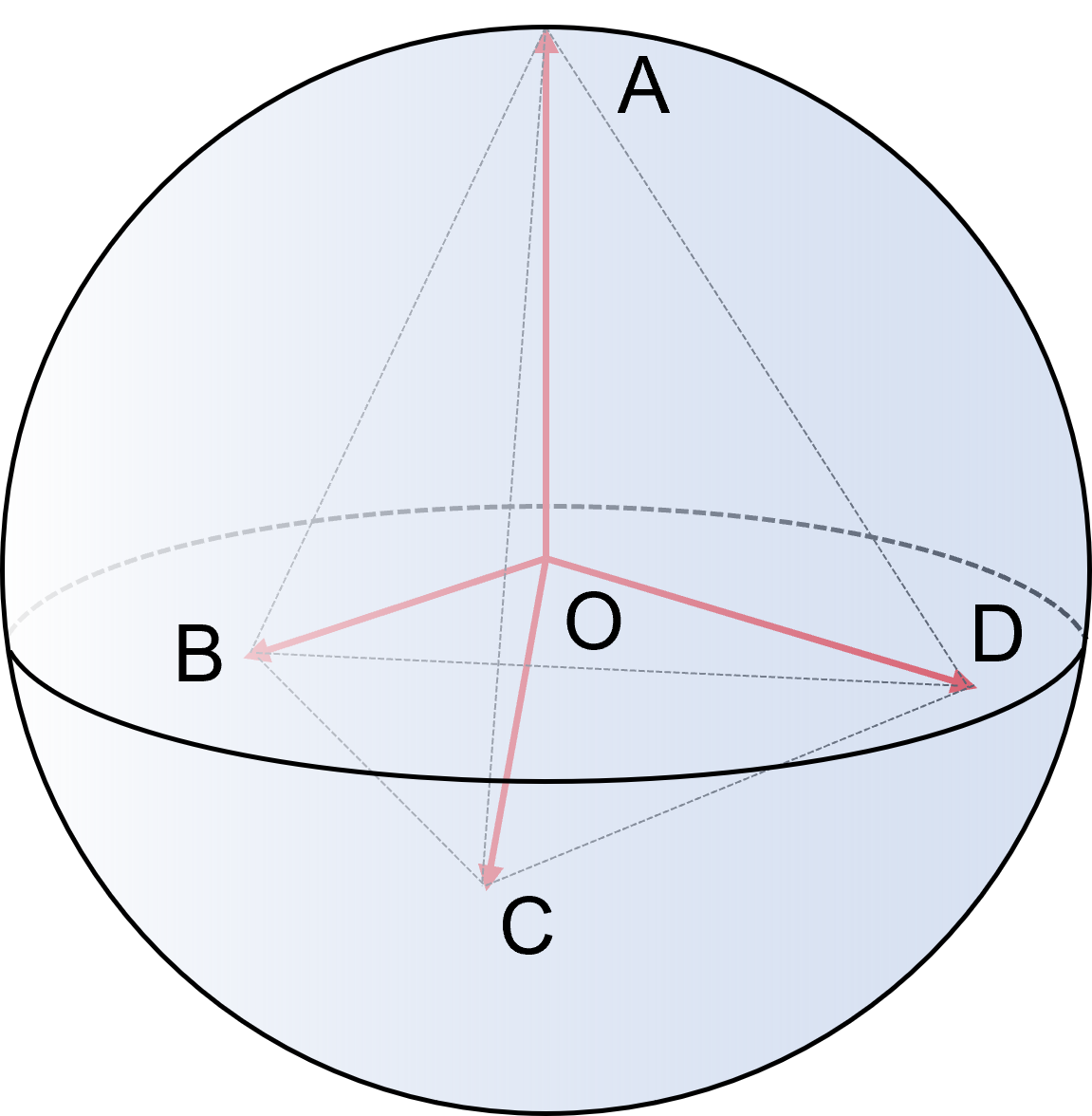} & 
    \includegraphics[height=1.1in]{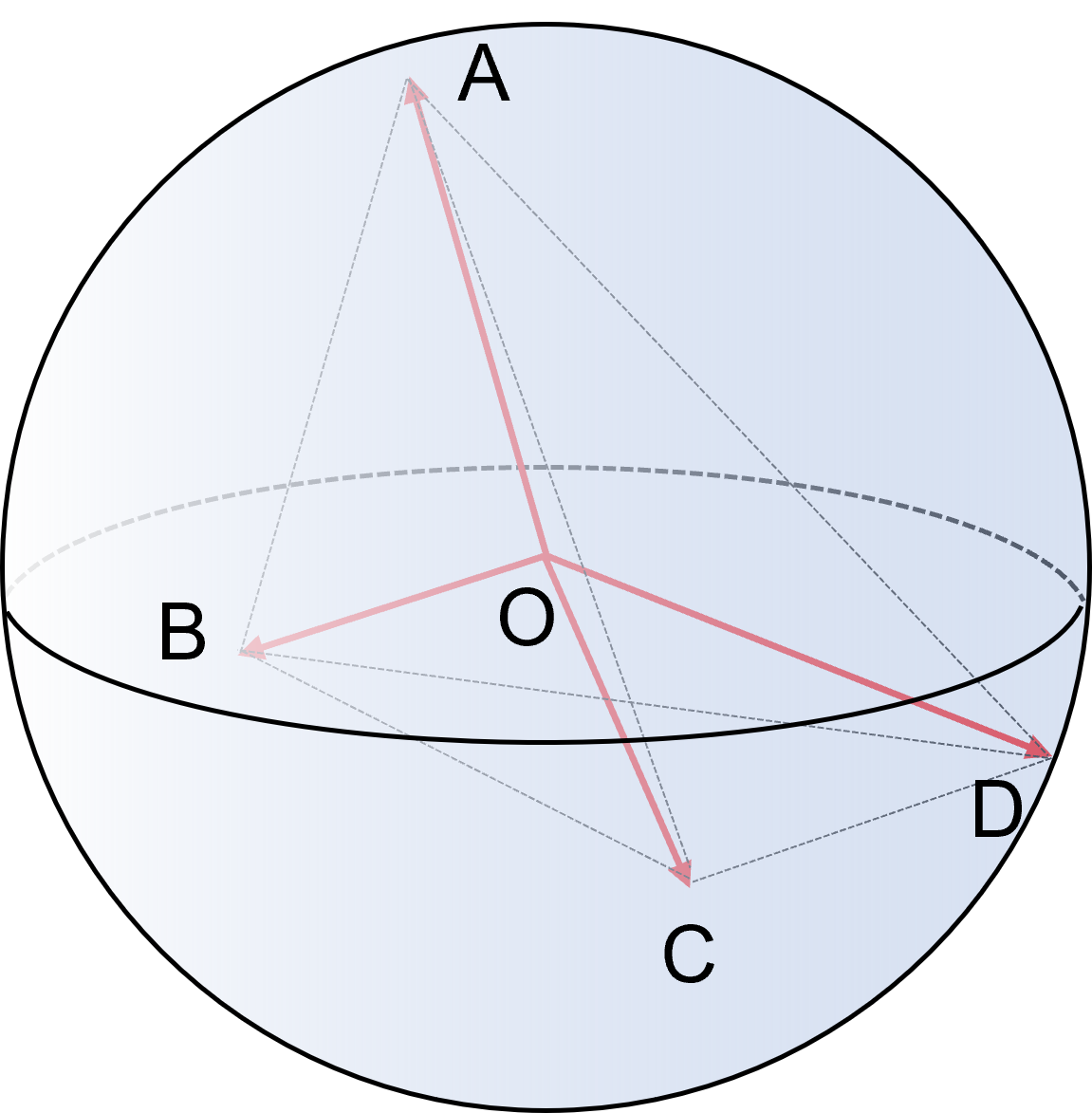} \\
    (a) & (b) \\
    & \\
    \includegraphics[height=0.85in]{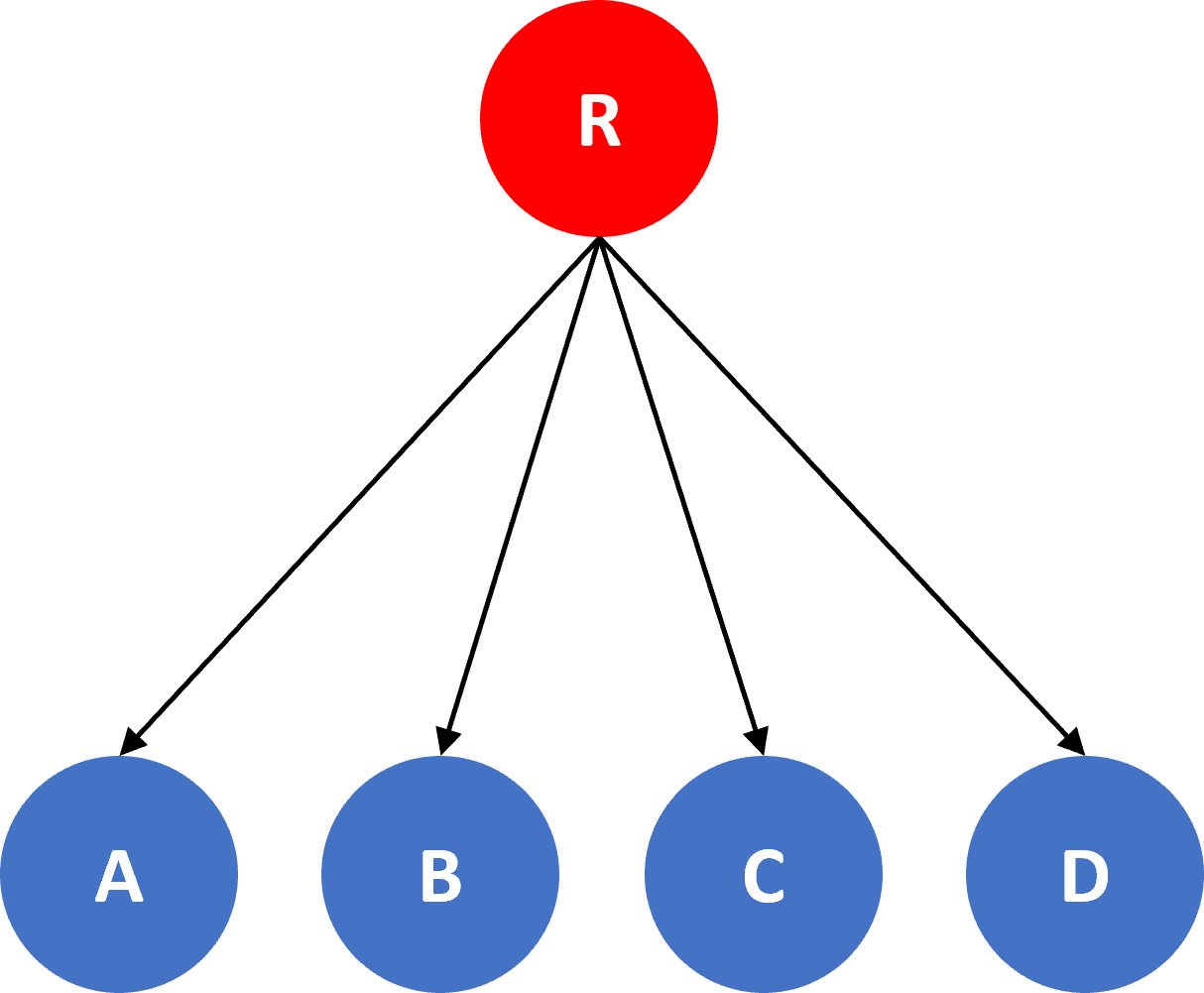} &
    \includegraphics[height=0.85in]{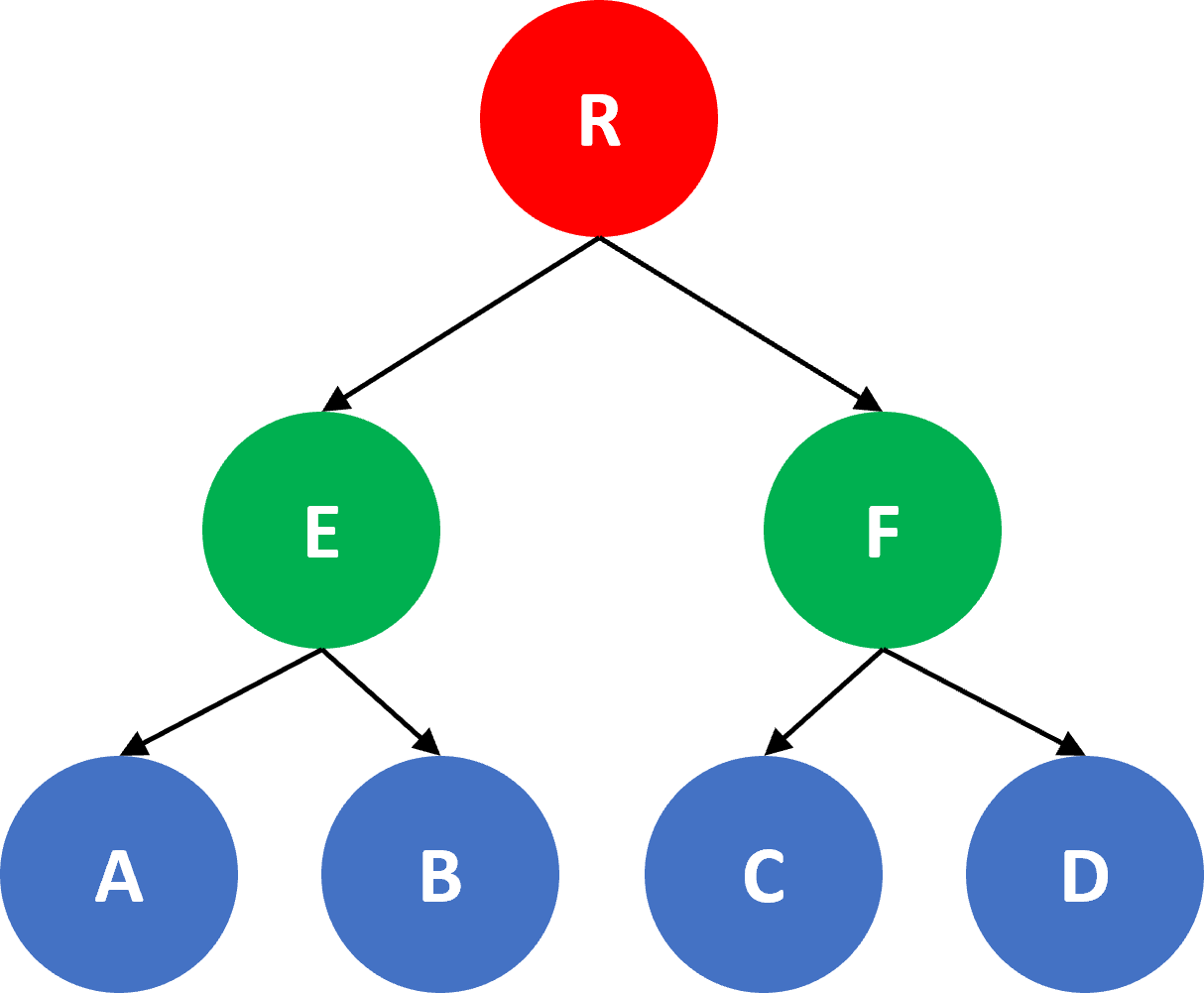} \\
    (c) & (d) \\
    \end{tabular}
    \caption{Illustration of a hierarchy-agnostic ETF (a) and a hierarchy-aware HAFrame (b) of four leaf classes and their hierarchies (c) and (d), respectively. All leaf classes in ETF have the same hierarchical distance.}
    \label{fig: illustrate}
\end{figure}

A recent study \cite{neuralcollapse} has unveiled a phenomenon termed neural collapse. It empirically revealed that the penultimate features of the same class tend to collapse to their within-class mean. The within-class means of all classes and their respective classifier weights tend to collapse to the vertices of a simplex Equiangular Tight Frame (ETF). A simplex ETF is a geometric structure that maximally separates the pair-wise angles of the $K$ vectors in $\mathbb{R}^d$, $d\geq K$, and the respective maximal pair-wise cosine similarity of these $K$ vectors is $\frac{-1}{K-1}$. As illustrated in Fig.~\ref{fig: illustrate}(a), when $K=4$, the simplex ETF reduces to a tetrahedron, and we can visualize this tetrahedron in 3D space via PCA projection since its geometry is 3D. One can view such ETF as an embedding of a hierarchy of four classes sharing the same root node as their parent. This hierarchy is visualized in Fig.~\ref{fig: illustrate}(c), where the four classes are equally separated regarding their hierarchical distance from each other. 

Intuitively, the neural collapse phenomenon makes sense considering an ETF separates all classes equally and maximally from each other. However, such a structure may not emerge when trained with imbalanced data. Features of minor classes may collapse to the same vector (minority collapse) \cite{layerpeeled}. Therefore, some studies encourage the features to form an ETF structure by fixing the classifier weights at a pre-computed ETF \cite{ETF_classifier} or employing additional regularizers to induce neural collapse \cite{ETF_regularizer, ETF_deep_long_tail}. In the context of reducing mistake severity, this raises another concern: when the ETF classifier makes a mistake, it is mainly random due to its equiangular nature. Similarly, conventional neural networks are trained mainly with cross-entropy and one-hot labels, ignoring any underlying hierarchical label relationships. The associated performance evaluations also focus on the top-1 accuracy of the predictions, treating all mistakes equally. In real-world application scenarios, some classification mistakes would have a much worse impact than others, e.g., mistaking a human for a tree in autonomous driving. Hence, it is critical to incorporate mistake severity into the performance evaluations and develop methods to reduce the mistake severity of the model predictions. One off-the-shelf way to define the severity of mistakes is by leveraging the hierarchical label relationship between incorrect predictions and their ground truths.

To impose a preferred error structure, we propose to fix the classifier vectors to a Hierarchy-Aware Frame (HAFrame) instead of an ETF to ensure that the classifier vectors of certain classes are ``closer" than others. Consequently, when a mistake occurs, it is more likely to fall onto a ``closer" class in the HAFrame, resulting in less mistake severity. An example of such a HAFrame is shown in Fig.~\ref{fig: illustrate}(b). Compared to the ETF for four classes, the HAFrame captures the pair-wise hierarchical distances across the four classes from a hierarchy shown in Fig.~\ref{fig: illustrate}(d), with class $A$ closer to $B$, and $A$ equally distant to $C$ and $D$. \\

\noindent 
There are several contributions of our work:
\begin{itemize}[noitemsep,nolistsep]
    \item Our approach is easy to adapt to different hierarchies as we only require a minimal change of the classifier and no architectural change of the backbone network. 
    \item Our approach provides an analytical solution to embed the hierarchical relationship of classes into the respective classifier.
    \item Our approach offers a new route to reduce mistake severity and the average hierarchical distance of predictions from the perspective of neural collapse. 
\end{itemize}

\section{Related Works} \label{sect:related_works}
\noindent

In recent years, multiple works have incorporated the semantic relationship of labels derived from text data or given by an explicit hierarchy to improve the classification of images or text. In this paper, we mainly discuss works closely related to incorporating hierarchical label relationships for image classification and reduction of mistake severity. 

\noindent
\textbf{Hierarchy-aware label methods}. These methods often utilize label embeddings to incorporate hierarchical label relationships into the model. 
In \cite{BetterMistakes_hie_loss}, soft-label embeddings derived from the hierarchical distances are proposed, and the KL-divergence from softmax scores to the soft-labels is minimized.  The label embeddings capturing pair-wise similarities of the classes are fixed on a unit hypersphere in \cite{Barz_Denzler}, or learned as hyperbolic embeddings in \cite{hyperbolicEmbed_Medical}, then the image features are induced to align with the label embeddings. Aside from deriving the label embeddings from an explicit hierarchy, there are also works \cite{DeVISE, ConSE, HyperbolicDeVISE, CLIP_model} that model hierarchical label relationships implicitly via learning the associated semantic embeddings of the labels from text data. These methods maximize the similarity between the visual embeddings learned from images and the corresponding semantic embeddings learned from text data.

\noindent
\textbf{Hierarchy-aware loss methods}.
A Hierarchy and Exclusion (HXE) graph is proposed in \cite{HXE_Graph} to model label relationships with a probabilistic classification model on the HXE graph capturing the semantic relationships (mutual exclusion, overlap, and subsumption) between any two labels. 
In \cite{BetterMistakes_hie_loss}, a hierarchical cross-entropy loss is proposed for the label tree. To integrate the knowledge of label relationships in a directed acyclic graph into the deep neural network, the probability of each label occurring is modeled independently to allow multi-label scenarios in \cite{Brust_Denzler}. In \cite{HieLossOptimalTransport}, the authors cast hierarchical classification as a discrete optimal transportation problem with an associated optimal transport loss. 

\begin{figure*}[t]
\centering
\setlength{\tabcolsep}{0.4pt}
\begin{tabular}{c c c c}

\includegraphics[height=1.4in]{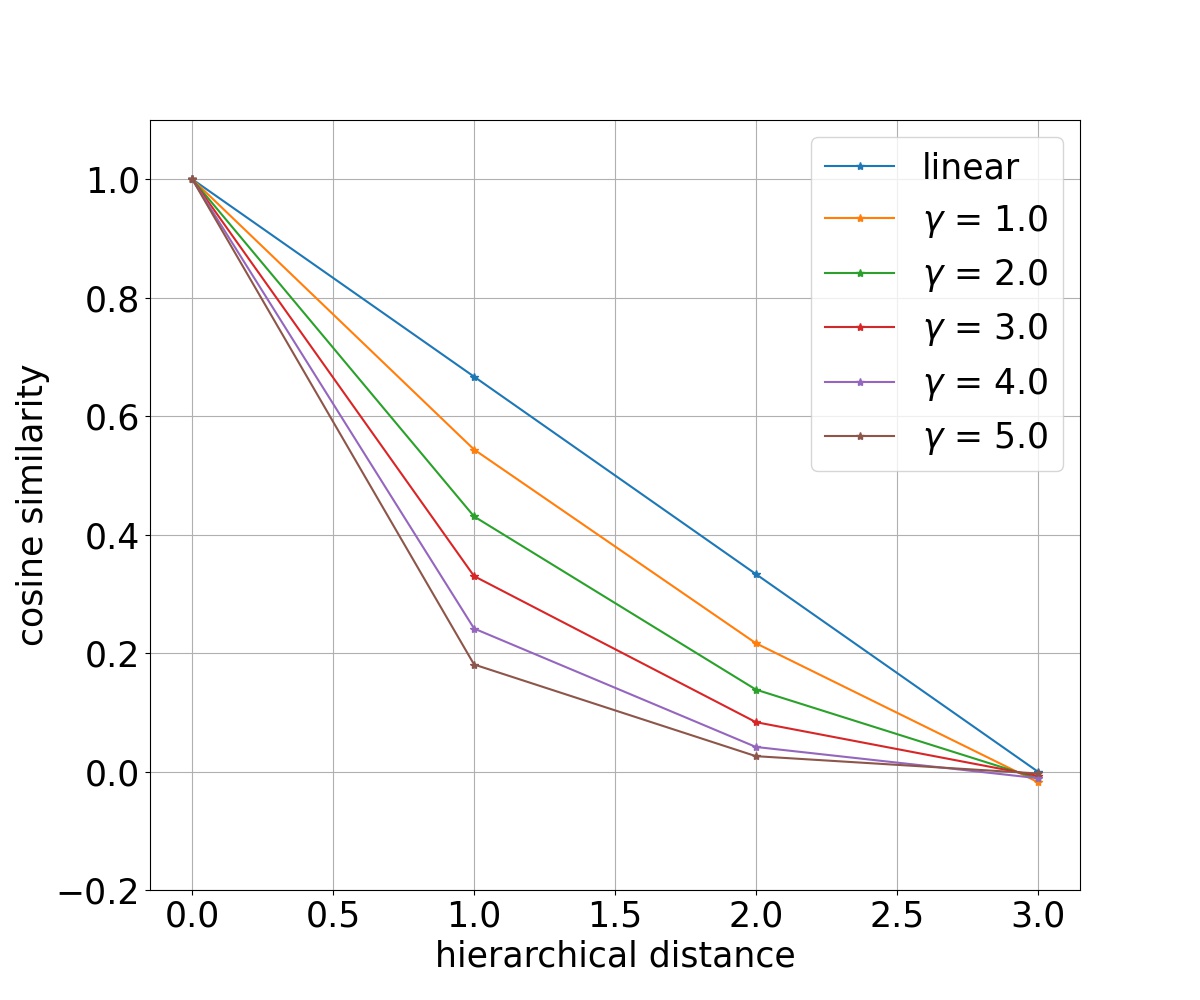} &
\includegraphics[height=1.4in]{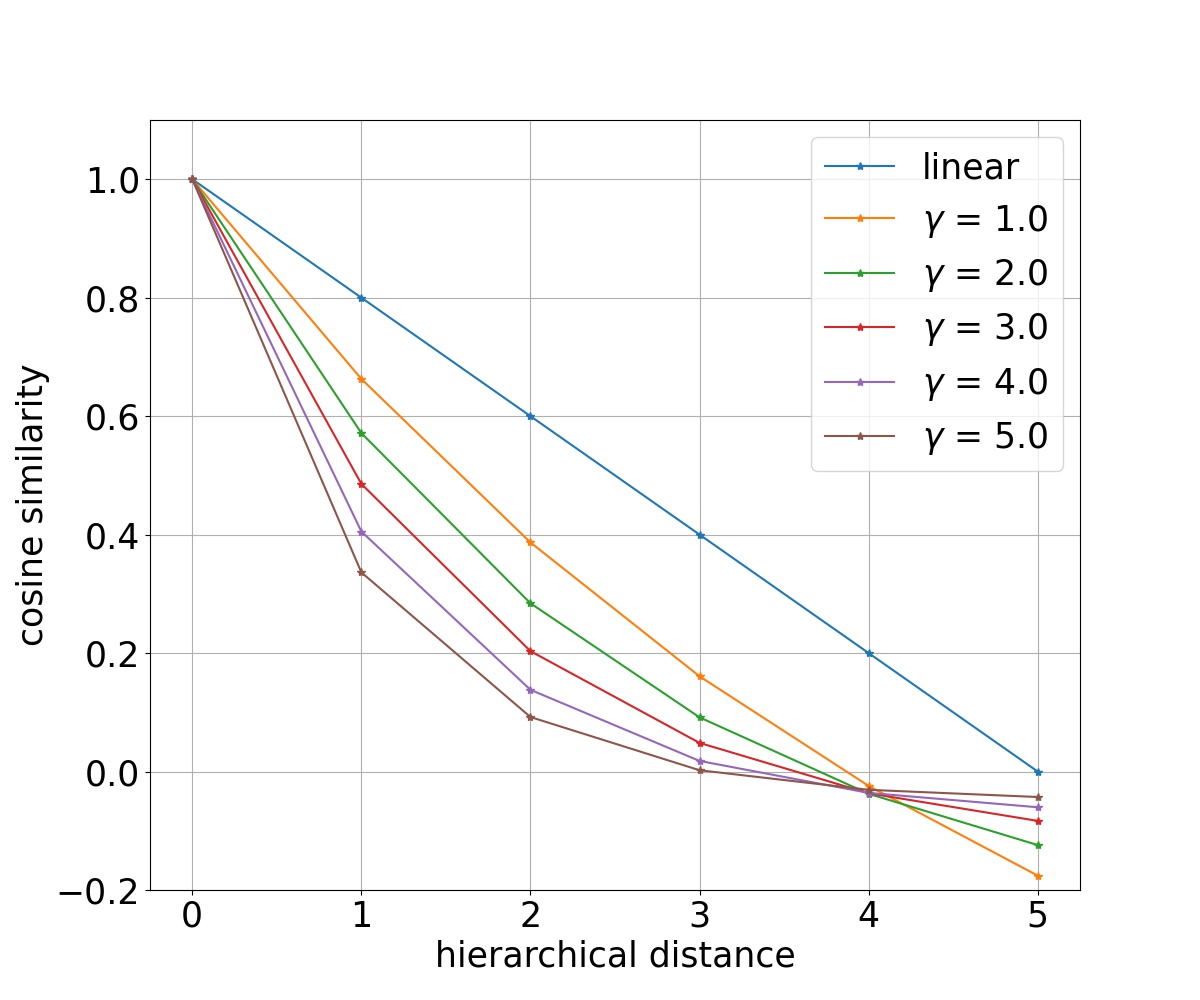} &
\includegraphics[height=1.4in]{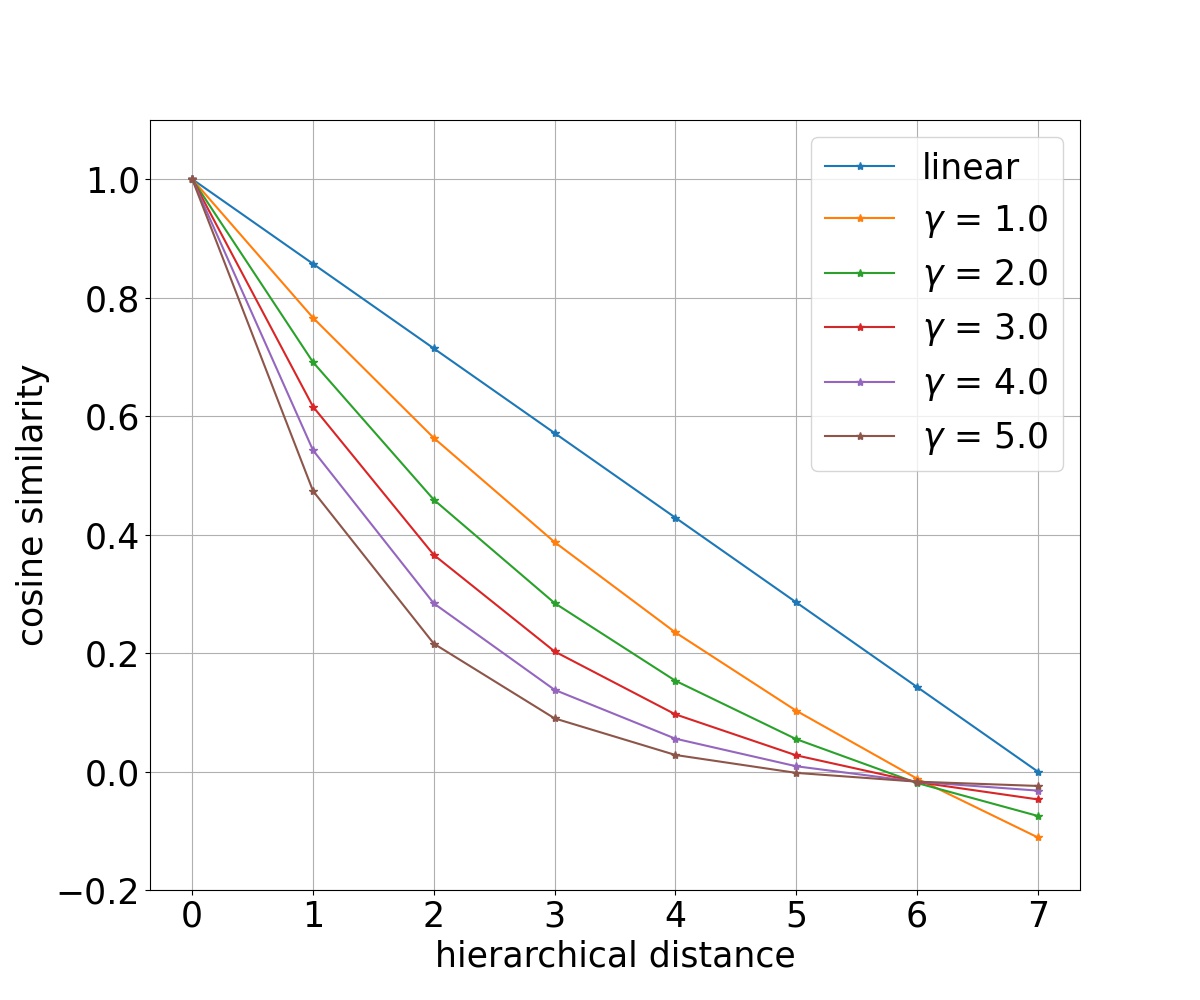} &
\includegraphics[height=1.4in]{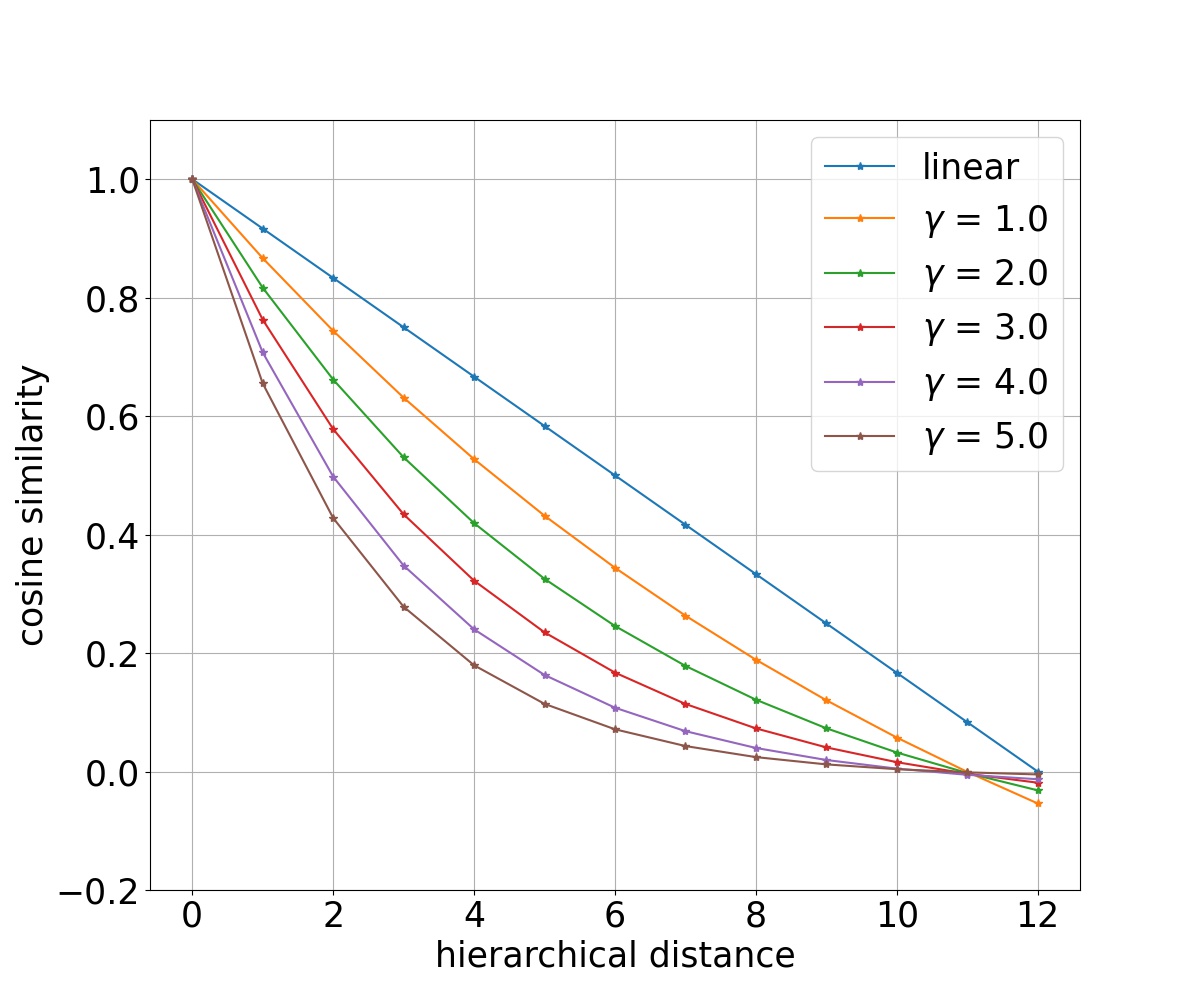} \\
(a) FGVC-Aircraft \cite{fgvc_aircraft} & (b) CIFAR-100 \cite{CIFAR} & (c) iNaturalist2019 \cite{iNat2019} & (d) tiered-ImageNet-H \cite{tieredimagenet}

\end{tabular}
\caption{Visualization of the exponential mapping functions given in Eqn.~\ref{eq: mapping_hie_to_cos}. The x-axis is hierarchical distance, and the y-axis is mapped cosine similarity. The linear mapping used in \cite{Barz_Denzler} is also plotted. The associated hierarchies used for each dataset are introduced in Sect.~\ref{sect: experiments}.} 
\label{fig: mappings}
\end{figure*}

\noindent
\textbf{Hierarchy-aware cost methods.} This line of research derives a cost measurement for misclassifications  from a given label hierarchy. The cost is then applied to amend flat predictions at inference time. In both \cite{BetterMistakes_CRM} and \cite{DengJiaCost}, the cost is defined as the height of the lowest common ancestor between the ground truth and incorrect prediction, i.e., the semantic level at which the misclassification occurs. The associated classification problem is then formulated as a conditional risk minimization (CRM) problem. Similarly, the cost is also formulated as the loss of label specificity in \cite{HedgingYourBets}, which optimizes the accuracy-specificity trade-offs in hierarchical classification for a given lower bound accuracy requirement. 

\noindent
\textbf{Hierarchy-aware architecture methods}. These methods require architectural changes to the network. A dynamic-structured network with a unifying hierarchy for classes of different datasets is proposed in \cite{DSSPN}. During training, the sub-graph of the related classes is dynamically activated, incorporating data from different datasets. In \cite{DRTC}, the authors proposed sharing the parameters of the non-leaf tree node classifiers and calibrating the posterior probability distribution of labels with a stochastic tree sampling method during training. The classifiers corresponding to nodes in a label hierarchy are constrained on hierarchically connected sphere manifolds in \cite{spherical_manifold} to regularize the model performance. In \cite{HieNoveltyDetect}, the classifier for each non-leaf node in the hierarchy is equipped with a virtual novel class to perform top-down classification with novelty detection. 

In \cite{MBM_wu}, the authors use independent classification heads for classes at different levels of the label hierarchy. All levels of classifiers share the same penultimate features, and the cross-entropy losses of all levels are optimized jointly. Similarly, a multi-classification head network is proposed in \cite{multihead} to incorporate hierarchical label relationships, but each classification head is staged at a different depth of the backbone network. In \cite{flamingo} (Flamingo), the authors proposed to use multiple classification heads for different levels of classes in the hierarchy, where the penultimate features are decoupled into different segments for the respective coarse and fine-grained classifiers. The multi-classification head setting is also adopted in \cite{HAFeature} (HAFeature). The classifiers of all levels share the same penultimate features from the backbone network. Additional geometric constraints are placed on the parent and child classifiers in the hierarchy. The hierarchical relationships of labels in adjacent levels are also enforced by minimizing the Jensen-Shannon Divergence \cite{JSDivergence} between predictions of the coarse-level classifier and the soft-labels reconstructed from predictions of the next fine-level classifier.

\section{Method}

Inspired by prior works which fix the classifier to polytope \cite{RePoNet, RePoNet_more}, Hadamard matrix \cite{fix_cls_hadamard}, and simplex ETF \cite{ETF_classifier}, we propose to fix the linear classifier to a HAFrame with the hierarchical relationship between the leaf classes embedded into their pair-wise cosine similarities. During training, we employ a weighted loss consisting of the cross-entropy loss and the proposed cosine similarity-based auxiliary loss to induce penultimate features collapsing onto the associated classifier vectors (HAFrame) to achieve the desired reduction of mistake severity. The hierarchies required in our work are constrained to label trees, same as recent works \cite{BetterMistakes_hie_loss, BetterMistakes_CRM, HAFeature}.

\subsection{Pair-wise Cosine Similarity}

We use the height of the lowest common ancestor (LCA) of two leaf classes $y_i$ and $y_j$ in the given hierarchy as the measurement of their hierarchical distance, as used in previous works \cite{Barz_Denzler, BetterMistakes_hie_loss, BetterMistakes_CRM}: 
\begin{equation}
    d_{ij} = height(LCA(y_i, y_j))
\end{equation}
where $i, j \in \{1,2,...,K\}$ and $K$ is the number of leaf classes in the hierarchy. 
We propose to map the pairwise hierarchical distance $d_{ij}$ to pairwise cosine similarity $S_{ij}$ between leaf classes $y_i$ and $y_j$ by an exponential mapping function:
\begin{equation} \label{eq: mapping_hie_to_cos}
    S_{ij} = (1-s_{min})\cdot e^{-\gamma\cdot\frac{d_{ij}}{d_{max}}} + s_{min}
\end{equation}
where $d_{max}$ is the height of the hierarchy, $\gamma >0$ is a hyper-parameter controlling the ``spacing" between hierarchically adjacent classes, and $s_{min}$ is a lower bound for the mapped cosine similarity: $S_{min} < S_{ij} \leq 1$. 

We can construct a real-valued symmetric cosine similarity matrix $\boldsymbol{S}$, where $\boldsymbol{S}_{i,j} = \boldsymbol{S}_{j,i} = S_{ij}$. For a given $\gamma$ and a set of hierarchical distances $d_{ij}, \forall i,j\in\{1,...,K\}$, we search the minimum $s_{min}$ between -1.0 and 1.0 with a step size of 0.02 such that the resulting $\boldsymbol{S}$ from our mapping (Eqn.~\ref{eq: mapping_hie_to_cos}) is \textit{positive definite}. Other mapping functions can also be used here as long as the associated similarity matrix $\boldsymbol{S}$ is guaranteed to be positive definite.

We plotted the linear mapping function $S_{ij} = 1 - d_{ij}/d_{max}$ used in \cite{Barz_Denzler} and our mapping function with $\gamma =1,2,...,5$ in Fig.~\ref{fig: mappings} over four datasets with increasing heights of the associated hierarchies for comparison. Our exponential mapping function is more flexible than the linear mapping function. The $\gamma$ parameter allows for a trade-off between separating hierarchically close and distant classes. In terms of the resulting cosine similarities, a larger $\gamma$ stretches hierarchically close classes further away while compressing hierarchically distant classes closer to each other.

\begin{figure*}[t]
\centering
\setlength{\tabcolsep}{0.4pt}

\includegraphics[height=1.5in]{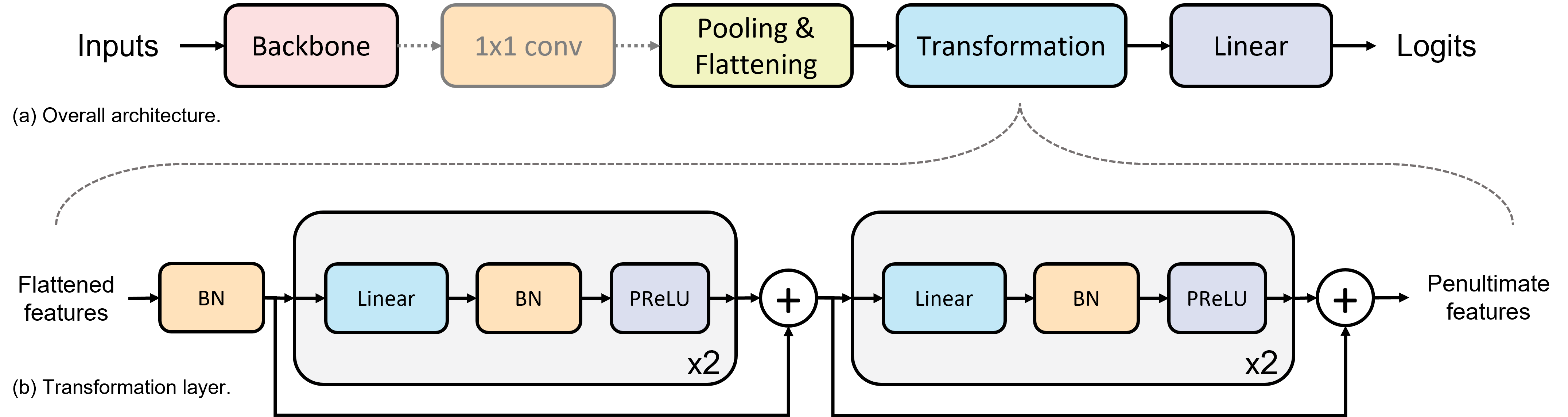}

\caption{Illustration of the customized network architectures. (a) Top: overall network architecture of our approach, the 1x1 convolutional layer is only used in our type-II models. (b) Bottom: the proposed transformation layer, where BN is 1D batch norm layer.}
\label{fig: architectures}
\end{figure*}

\subsection{Hierarchy-Aware Frame}

In this section, we introduce the proposed HAFrame and how to solve it from the positive definite pair-wise cosine similarity matrix $\boldsymbol{S}$ derived from a given hierarchy.

\begin{definition}[Hierarchy-Aware Frame]
Let a set of vectors $\{\boldsymbol{w}_i\}_{i=1}^K$ in $\mathbb{R}^K$ with $||\boldsymbol{w}_i||_2=1$, $i=1,2,...,K$, and their pair-wise cosine similarities satisfy the following equation:
\begin{equation}
    cos\angle(\boldsymbol{w}_i, \boldsymbol{w}_j)=\boldsymbol{w}^T_i \boldsymbol{w}_j=S_{ij} , \forall {1\leq i\leq j\leq K} 
\end{equation}
where $S_{ij}$ is the cosine similarity between classes $i$ and $j$ given in Eqn.~\ref{eq: mapping_hie_to_cos}. Next, let $\boldsymbol{W}=[\boldsymbol{w}_1\:\boldsymbol{w}_2\:...\:\boldsymbol{w}_K]$ be the proposed hierarchy-aware frame, $\boldsymbol{W}$ is in $\mathbb{R}^{K\times K}$ and satisfies
\begin{equation} \label{eqn: s=wTw}
    \boldsymbol{S} = \boldsymbol{W}^T\boldsymbol{W}
\end{equation}
Since $\boldsymbol{S}$ is guaranteed to be positive definite by our search for the proper $s_{min}$ in the mapping function (Eqn.~\ref{eq: mapping_hie_to_cos}), we can find $\boldsymbol{W}$ by the following matrix factorization:
\begin{equation} \label{eq:factorization}
    \boldsymbol{S} = \boldsymbol{Q}\boldsymbol{D}\boldsymbol{Q}^T = 
    (\boldsymbol{Q}\boldsymbol{D}^{\frac{1}{2}}\boldsymbol{U}^T)(\boldsymbol{U}\boldsymbol{D}^{\frac{1}{2}}\boldsymbol{Q}^T)=\boldsymbol{W}^T\boldsymbol{W}
\end{equation}
where $\boldsymbol{Q} \in \mathbb{R}^{K \times K}$ and $\boldsymbol{D}\in\mathbb{R}^{K\times K}$ are acquired from eigenvalue decomposition of $\boldsymbol{S}$, and $\boldsymbol{U}\in \mathbb{R}^{K\times K}$ is an orthonormal matrix obtained from QR-decomposition of a randomly sampled matrix in $\mathbb{R}^{K\times K}$ that allows an arbitrary rotation and satisfies $\boldsymbol{U}^T\boldsymbol{U}=\boldsymbol{U}\boldsymbol{U}^T=\boldsymbol{I}_K$. Therefore, the proposed hierarchy-aware frame is given by:
\begin{equation} \label{eq: HAF_solution}
    \boldsymbol{W} = \boldsymbol{U}\boldsymbol{D}^{\frac{1}{2}}\boldsymbol{Q}^T
\end{equation}
\end{definition}
We prove that $\boldsymbol{W}$ (i.e., our HAFrame) satisfies the \textit{frame condition} \cite{frameCond} as long as the similarity matrix $\boldsymbol{S}$ is positive definite in the supplementary material.

\subsection{Additional Transformation Layer} \label{sect: transformation}

Since we are encouraging the penultimate features $\boldsymbol{h}$, instead of the centered features $\tilde{\boldsymbol{h}}$ ($\tilde{\boldsymbol{h}} = \boldsymbol{h}-\boldsymbol{h}_G$, where $\boldsymbol{h}_G$ is the average of all penultimate features) described in \cite{neuralcollapse}, to collapse onto their respective classifier vectors, we propose to use an additional transformation layer before the final classification layer to facilitate such collapse. The overall architecture of our model is shown in Fig.~\ref{fig: architectures}(a). 

Our transformation layer needs to learn a mapping of the features to classifiers potentially residing in different quadrants (including the negative quadrant) of the Euclidean space $\mathbb{R}^K$. The entries of penultimate feature from most common convolutional neural networks \cite{ResNet, EfficientNet, InceptionV3} in recent years have an inductive bias towards non-negative values due to the use of nonlinear activation functions (e.g., ReLU \cite{ReLU}, GELU \cite{GELU}, etc.). Our proposed transformation layer uses parametric ReLU (PReLU) \cite{PReLU}, which learns the slope of the rectified linear function for negative inputs to mitigate the aforementioned bias, therefore enabling the transformed features to have negative entries approximating the respective fixed classifier vectors of $\boldsymbol{W}$. We also employ residual connections to improve information flow during training. The architecture of the transformation module is shown in Fig.~\ref{fig: architectures}(b). All four linear layers in the transformation module have $K$ (number of classes) hidden units as the classifier vectors $\{\boldsymbol{w}_i\}_{i=1}^K$ of HAFrame are in $\mathbb{R}^K$. To accommodate this requirement, we add a 1x1 convolutional layer between the backbone and the pooling layer to reduce the number of channels in the backbone features to $K$.

\subsection{Cosine Similarity Based Auxiliary Loss}
Once we solve the hierarchy-aware frame $\boldsymbol{W}$ for a given pair-wise similarity matrix $\boldsymbol{S}$, the associated column vectors in $\boldsymbol{W}$ are used as weight vectors in the linear classification layer of the network. The corresponding bias term for the linear layer is removed. The prediction $\hat{y}$ is given by:
\begin{equation}
    \hat{y} = \argmax_i \boldsymbol{W}^T\boldsymbol{h} 
            = \argmax_i cos\angle(\boldsymbol{w}_i, \boldsymbol{h})
\end{equation}
where $\boldsymbol{h} \in \mathbb{R}^{K}$ is the penultimate feature vector of an input example produced by the transformation layer. The logits of an example are given by $\boldsymbol{W}^T\boldsymbol{h}$, consistent with a regular linear classifier omitting the bias term. In addition to cross-entropy loss, we propose to use a cosine similarity-based auxiliary loss to facilitate the collapse of penultimate features onto the respective classifier weights:
\begin{equation} \label{loss: cosine}
    \mathcal{L}_{COS} = \sum_{i=1}^{K} ( cos\angle(\boldsymbol{w}_i,\boldsymbol{h}) - cos\angle(\boldsymbol{w}_i,\boldsymbol{w}_y) )^2
\end{equation}
where $y$ is the ground truth of $\boldsymbol{h}$ and $\boldsymbol{w}_y$ is the fixed classifier vector corresponding to class $y$. This auxiliary loss reduces to zero when $cos\angle(\boldsymbol{h},\boldsymbol{w}_y)=1$, i.e., the penultimate feature vector $\boldsymbol{h}$ is aligned to the same direction of its classifier $\boldsymbol{w}_y$.

The overall training loss is a mixture of cross-entropy loss ($\mathcal{L}_{CE}$) and the proposed auxiliary loss with a hyperparameter $\alpha \in [0, 1]$ controlling the mixing ratio:
\begin{equation} \label{eq: total_loss}
    \mathcal{L} = (1-\alpha)\mathcal{L}_{CE} + \alpha \mathcal{L}_{COS}
\end{equation}
This loss is averaged across all training examples in a mini-batch.

\section{Experiments} \label{sect: experiments}

We compare our approach with a baseline model trained with cross-entropy and recent competitive methods of CRM \cite{BetterMistakes_CRM}, Flamingo \cite{flamingo}, and HAFeature \cite{HAFeature} introduced in Sect.~\ref{sect:related_works}, we omit comparison with earlier approaches \cite{DeVISE, Barz_Denzler, BetterMistakes_hie_loss} as more recent works \cite{BetterMistakes_CRM, HAFeature} have demonstrated that the results of earlier methods are suboptimal. Since CRM is an inference-time approach to modifying flat predictions, we apply CRM to the predictions of the baseline model.

\noindent
\textbf{Datasets.} Following the previous works \cite{BetterMistakes_hie_loss, BetterMistakes_CRM}, we conduct comparative experiments on the tieredImageNet-H \cite{tieredimagenet, BetterMistakes_hie_loss} and iNaturalist2019 \cite{iNat2019} datasets. We also include comparison results on CIFAR-100 \cite{CIFAR} and FGVC-Aircraft \cite{fgvc_aircraft}. The four datasets used in our experiments scale from 100 classes to 1010 classes, and their respective hierarchy's height ranges from 3 to 12. For FGVC-Aircraft, we adopt the original hierarchy provided by the dataset. For CIFAR-100, we use the hierarchy provided by the Flamingo approach \cite{flamingo}. As for iNaturalist2019 and tieredImageNet-H, we adopt their respective hierarchies from \cite{BetterMistakes_hie_loss}. The statistics of these datasets are summarized in Table~\ref{tab:dataset_stats}.

\begin{table}[h]
\centering
\small\addtolength{\tabcolsep}{-3pt}
\begin{tabular}{c|c c c c c}
    
 \hline   
 Dataset          & Height  & Classes  &  Train   & Val       & Test   \\
 \hline
 FGVC-Aircraft    & 3        & 100      & 3,334    & 3,333     & 3,333  \\
 CIFAR-100        & 5        & 100      & 45,000   & 5,000     & 10,000 \\
 iNaturalist2019   & 7        & 1010     & 187,385  & 40,121    & 40,737 \\
 tieredImageNet-H & 12       & 608      & 425,600  & 15,200    & 15,200 \\
\hline
\end{tabular}
\caption{Statistics of the four datasets used in our experiments.}
\label{tab:dataset_stats}
\end{table}

\noindent
\textbf{Training Configurations.} \label{subsect: modeltypes}
We use ResNet-50 \cite{ResNet} as the backbone network to evaluate all methods on FGVC-Aircraft, iNaturalist2019, and tieredImageNet-H. For CIFAR-100, we adopt a WideResNet-28 \cite{WideResNet} backbone for all methods, following experimental settings in HAFeature \cite{HAFeature}. We initialize all models on FGVC-Aircraft and iNaturalist2019 with ImageNet-1K \cite{Imagenet} pretrained weights. The WideResNet models are trained on CIFAR-100 without pretrained weights. The previous works \cite{BetterMistakes_hie_loss, BetterMistakes_CRM, HAFeature} all used pretrained weights from ImageNet-1K, a superset of tieredImageNet-H, to initialize models on tieredImageNet-H. We instead use pretrained weights from the PASS dataset \cite{PASS} (i.e., an ImageNet replacement for self-supervised pretraining without humans) for all models on tieredImageNet-H to avoid refitting the model to the same data already seen by its pretrained weights. 

We evaluate all methods with two penultimate feature extraction models for fair and sufficient comparison. The first model (type-I) follows the Flamingo/HAFeature \cite{flamingo, HAFeature} settings by adding an extra transformation layer before the classification layer, similar to our approach shown in Fig.~\ref{fig: architectures}(a) but without the 1x1 convolutional layer before the pooling layer. This extra transformation layer consists of a linear layer with 600 hidden units and two batch norm layers before and after the linear layer. The output is rectified with ELU \cite{ELU} activation. The second model (type-II, our architecture) adds a 1x1 convolutional layer before the pooling layer and replaces the type-I model's transformation layer with the proposed transformation layer introduced in Sect.~\ref{sect: transformation}.

The models across all datasets are trained for 100 epochs. The type-I models follow the training strategy of the Flamingo approach \cite{flamingo}, i.e., using SGD optimizer with 0.01 as the initial learning rate (LR) for the backbone network and 0.1 as the initial LR for both transformation and the classification layers. The type-II models (using the proposed transformation layer) use 0.01 as the initial LR for both the backbone and transformation layer and 0.1 as the initial LR for the classification layers. We make such changes to type-II models as this yields better results. Both type-I and type-II models are trained with a cosine annealing learning rate scheduler used by the Flamingo approach \cite{flamingo}. The batch size of all models trained on iNaturalist2019 and tieredImageNet-H is 256. We find that a smaller batch size of 64 yields overall better results for models on CIFAR-100 and FGVC-Aircraft. During training, we select the model with the highest validation accuracy for subsequent evaluation. 

\begin{table*}[t]
\centering
\footnotesize\addtolength{\tabcolsep}{1.5pt}
\begin{tabular}{c|c|c c c c c}
 \hline   
  Model & Method  &  Top-1 Accuracy $\uparrow$ & Mistake Severity $\downarrow$ & HierDist@1 $\downarrow$ & HierDist@5 $\downarrow$ & Hierdist@20 $\downarrow$ \\
 \hline
 \multirow{4}{*}{Type-I} & cross-entropy   & 79.18 +/- 0.5511  & 2.12 +/- 0.0240  & 0.44 +/- 0.0097  & 2.10 +/- 0.0033  & 2.67 +/- 0.0040 \\
 
 & CRM \cite{BetterMistakes_CRM}             & 79.30 +/- 0.5250  & 2.08 +/- 0.0201  & 0.43 +/- 0.0091  & 1.74 +/- 0.0040  & 2.44 +/- 0.0015 \\
 
 & \cellcolor{blue!15}Flamingo \cite{flamingo}        & \cellcolor{blue!35} 81.00 +/- 0.5873  & \cellcolor{blue!15}2.04 +/- 0.0343  & \cellcolor{blue!15}0.39 +/- 0.0072  & \cellcolor{blue!15}2.06 +/- 0.0041  & \cellcolor{blue!15}2.65 +/- 0.0018 \\
 
 & HAFeature \cite{HAFeature}      & 73.23 +/- 0.6085  & 2.48 +/- 0.0937  & 0.66 +/- 0.0152  & 2.10 +/- 0.0126  & 2.61 +/- 0.0078 \\
 \hline
\multirow{5}{*}{Type-II} & cross-entropy   & 79.58 +/- 0.2727 & 2.15 +/- 0.0159 & 0.44 +/- 0.0067 & 2.11 +/- 0.0055 & 2.67 +/- 0.0034 \\
 & CRM \cite{BetterMistakes_CRM}           & 79.62 +/- 0.2953 & 2.13 +/- 0.0109 & 0.43 +/- 0.0058 & 1.75 +/- 0.0043 & 2.45 +/- 0.0022 \\
 & Flamingo \cite{flamingo}      & 80.02 +/- 0.7886 & 2.10 +/- 0.0373 & 0.42 +/- 0.0168 & 2.08 +/- 0.0058 & 2.66 +/- 0.0031 \\
 & HAFeature \cite{HAFeature}      & 74.39 +/- 0.7813 & 2.53 +/- 0.0334 & 0.65 +/- 0.0226 & 2.10 +/- 0.0064 & 2.61 +/- 0.0041 \\
 & \cellcolor{blue!15}HAFrame (ours) & \cellcolor{blue!15}80.49 +/- 0.4692 & \cellcolor{blue!35} 2.02 +/- 0.0381 & \cellcolor{blue!35} 0.39 +/- 0.0039 & \cellcolor{blue!35} 1.74 +/- 0.0027 & \cellcolor{blue!35} 2.45 +/- 0.0024 \\
\hline
\end{tabular}
\caption{Experiment results on FGVC-Aircraft dataset. The details of type-I and type-II models are included in the training config.} 
\label{tab:fgvc_aircraft}
\end{table*}

\begin{table*}[t]
\centering
\footnotesize\addtolength{\tabcolsep}{1.5pt}
\begin{tabular}{c|c|c c c c c}
 \hline   
 Model & Method          &  Top-1 Accuracy $\uparrow$ & Mistake Severity $\downarrow$ & HierDist@1 $\downarrow$ & HierDist@5 $\downarrow$ & Hierdist@20 $\downarrow$ \\
 \hline
 \multirow{4}{*}{Type-I} & cross-entropy   & 77.65 +/- 0.2635  & 2.34 +/- 0.0271  & 0.52 +/- 0.0102  & 2.25 +/- 0.0084 & 3.19 +/- 0.0045 \\
 
 & CRM \cite{BetterMistakes_CRM}            & 77.63 +/- 0.2800  & 2.30 +/- 0.0255  & 0.51 +/- 0.0093  & 1.11 +/- 0.0077 & 2.18 +/- 0.0028 \\
 
 & Flamingo \cite{flamingo}        & 77.91 +/- 0.5733  & 2.31 +/- 0.0179  & 0.51 +/- 0.0137  & 2.07 +/- 0.0198 & 3.08 +/- 0.0094 \\
 
 & \cellcolor{blue!15}HAFeature \cite{HAFeature}       & \cellcolor{blue!15}77.49 +/- 0.3391  & \cellcolor{blue!15}2.24 +/- 0.0158  &\cellcolor{blue!15}0.51 +/- 0.0084  &\cellcolor{blue!15} 1.43 +/- 0.0108 &\cellcolor{blue!15} 2.64 +/- 0.0105 \\
 
 \hline
\multirow{5}{*}{Type-II} & cross-entropy  & 76.45 +/- 0.2207  & 2.43 +/- 0.0235  & 0.57 +/- 0.0106  & 2.35 +/- 0.0049  & 3.30 +/- 0.0030 \\

& CRM \cite{BetterMistakes_CRM}           & 76.48 +/- 0.2278  & 2.38 +/- 0.0175  & 0.56 +/- 0.0095  & 1.15 +/- 0.0074  & 2.20 +/- 0.0029 \\

& Flamingo \cite{flamingo}      & 75.19 +/- 0.3188  & 2.31 +/- 0.0270  & 0.57 +/- 0.0043  & 2.42 +/- 0.0161  & 3.29 +/- 0.0105 \\      

& \cellcolor{blue!15}HAFeature  \cite{HAFeature}    &\cellcolor{blue!15} 76.44 +/- 0.1560  &\cellcolor{blue!15} 2.26 +/- 0.0290  &\cellcolor{blue!15} 0.53 +/- 0.0055  &\cellcolor{blue!15} 1.71 +/- 0.0130  &\cellcolor{blue!15} 2.84 +/- 0.0143 \\

& \cellcolor{blue!15}HAFrame (ours) &\cellcolor{blue!35} 77.71 +/- 0.2319  &\cellcolor{blue!35} 2.21 +/- 0.0108  &\cellcolor{blue!35} 0.49 +/- 0.0066  &\cellcolor{blue!35} 1.11 +/- 0.0018  &\cellcolor{blue!35} 2.18 +/- 0.0013 \\
\hline
\end{tabular}
\caption{Experiment results on CIFAR-100 dataset. The details of type-I and type-II models are included in the training config.} 
\label{tab:cifar100_fixed_train_val}
\end{table*}

\noindent
\textbf{Evaluation Metrics.} We adopt the evaluation metrics used in previous works \cite{BetterMistakes_hie_loss, BetterMistakes_CRM, HAFeature}: (1) top-1 accuracy; (2) average mistake severity, i.e., the sum of the heights of LCA between incorrect predictions and the respective ground truth labels averaged across all incorrect predictions; (3) average hierarchical distance at $k$, for $k=1,5,20$, i.e., HierDist@1, HierDist@5, HierDist@20, respectively. For each test example, the average height of LCA between the ground truth label and $k$ most likely predictions are evaluated, and the resulting average height for each test example is further averaged across the entire test set. We train five models for each method and report every evaluation metric's mean and 95\% confidence interval derived from the t-distribution with four degrees of freedom.

\noindent
\textbf{Hyperparameter Search.} We conducted a search for the proper $\gamma$ in the mapping function (Eqn.~\ref{eq: mapping_hie_to_cos}) and the mixing ratio $\alpha$ between cross-entropy and our cosine similarity auxiliary loss in Eqn.~\ref{eq: total_loss} using the validation examples of each dataset. For FGVC-Aircraft and CIFAR-100, we conducted a grid search of $\gamma$ and $\alpha$ on $\{1.0, 2.0, 3.0, 4.0, 5.0\} \times \{0.3, 0.4, 0.5, 0.6\}$.  For iNaturalist2019 and tieredImageNet, we searched on a smaller grid $\{3.0, 4.0, 5.0\} \times \{0.3, 0.4, 0.5\}$. Each configuration of $\gamma$ and $\alpha$ is evaluated on the validation set with three runs. We select the configuration that yields the best average hierarchical distance at 1 (HierDist@1) averaged across three runs. The resulting configurations ($\alpha$, $\gamma$) for FGVC-Aircraft, CIFAR-100, iNaturalist2019, and tieredImageNet-H are (0.5, 1.0), (0.4, 2.0), (0.4, 5.0), and (0.5, 3.0), respectively. 

\subsection{Results}

The experiment results on FGVC-Aircraft, CIFAR-100, iNaturalist2019, and tieredImageNet-H are shown in Tables~\ref{tab:fgvc_aircraft}, \ref{tab:cifar100_fixed_train_val}, \ref{tab:iNat2019}, and \ref{tab:tiered_imagenet}, respectively. The rows in the tables highlighted with \colorbox{blue!15}{light purple} are competitive methods. The method with the smallest (best) average mistake severity is selected first. Other methods with an average mistake severity not greater than the smallest mistake severity of 0.05 are also deemed competitive. Among these competitive methods, we highlight the best-performing entry for each metric with \colorbox{blue!35}{purple}. 

Our approach (HAFrame) has reached the best average mistake severity across all four datasets and the best top-1 accuracy on three datasets with a $\sim$0.51\% (80.49\%) drop of top-1 accuracy compared to the Flamingo approach (81.00\%) on FGVC-Aircraft (Table~\ref{tab:fgvc_aircraft}). Our approach also performs best on the average hierarchical distance among the competitive methods on three datasets. On tieredImageNet-H (Table~\ref{tab:tiered_imagenet}), our HierDist@5 and HierDist@20 are worse than, yet close to, CRM but still outperform Flamingo and HAFeature by a large margin. It is worth noting that CRM only reaches competitive average mistake severity on tieredImageNet-H (Table~\ref{tab:tiered_imagenet}), but its associated average hierarchical distances at $k=5$ and $k=20$ remain competitive or best on all datasets. The HAFeature approach improves average mistake severity better than CRM on three datasets but does not perform well on FGVC-Aircraft (Table~\ref{tab:fgvc_aircraft}) with a shallow hierarchy of 4 levels (including the root) and does not rank predictions of less likely classes well, i.e., its HierDist@5 and HierDist@20 do not perform as good as CRM or our approach. The average mistake severity of the Flamingo approach reaches competitive results on FGVC-Aircraft and tieredImageNet-H, but its average hierarchical distances exhibit suboptimal performance similar to the HAFeature approach. 

\begin{table*}[t]
\centering
\footnotesize\addtolength{\tabcolsep}{1.5pt}
\begin{tabular}{c|c|c c c c c}
    
 \hline   
 Model & Method          &  Top-1 Accuracy $\uparrow$ & Mistake Severity $\downarrow$ & HierDist@1 $\downarrow$ & HierDist@5 $\downarrow$ & Hierdist@20 $\downarrow$ \\
 \hline
 \multirow{4}{*}{Type-I} & cross-entropy   & 70.68 +/- 0.2097  & 2.22 +/- 0.0103  & 0.65 +/- 0.0068  & 1.95 +/- 0.0043  & 3.37 +/- 0.0040 \\
 & CRM \cite{BetterMistakes_CRM}            & 70.67 +/- 0.2095  & 2.16 +/- 0.0045  & 0.63 +/- 0.0057  & 1.17 +/- 0.0042  & 1.75 +/- 0.0033 \\
 
 & Flamingo \cite{flamingo}       & 70.11 +/- 0.1119  & 2.13 +/- 0.0063  & 0.64 +/- 0.0014  & 1.79 +/- 0.0126  & 3.28 +/- 0.0114 \\
 
 & HAFeature \cite{HAFeature}      & 70.57 +/- 0.1645  & 2.13 +/- 0.0192  & 0.63 +/- 0.0045  & 1.55 +/- 0.2188  & 2.68 +/- 0.4208 \\
 \hline
 \multirow{5}{*}{Type-II} & cross-entropy  & 70.44 +/- 0.1576  & 2.26 +/- 0.0071  & 0.67 +/- 0.0036  & 1.97 +/- 0.0060  & 3.40 +/- 0.0070 \\
 & CRM \cite{BetterMistakes_CRM} & 70.47 +/- 0.1363  & 2.21 +/- 0.0099  & 0.65 +/- 0.0029  & 1.18 +/- 0.0020  & 1.76 +/- 0.0016 \\
 
 & Flamingo \cite{flamingo}  & 70.13 +/- 0.1499  & 2.15 +/- 0.0061  & 0.64 +/- 0.0045  & 1.76 +/- 0.0037  & 3.31 +/- 0.0071 \\      
 
 & HAFeature \cite{HAFeature} &  68.46 +/- 4.6278 & 2.21 +/- 0.1298  & 0.70 +/- 0.1501  & 1.50 +/- 0.1235  & 2.49 +/- 0.0842 \\
 
 & \cellcolor{blue!15}HAFrame (ours)  & \cellcolor{blue!35}70.89 +/- 0.1213  & \cellcolor{blue!35}2.04 +/- 0.0107  & \cellcolor{blue!35}0.59 +/- 0.0033   & \cellcolor{blue!35}1.14 +/- 0.0033 & \cellcolor{blue!35}1.73 +/- 0.0023 \\
 
\hline
\end{tabular}
\caption{Experiment results on iNaturalist2019 dataset. The details of type-I and type-II models are included in the training config.}
\label{tab:iNat2019}
\end{table*}

\begin{table*}[t]
\centering
\footnotesize\addtolength{\tabcolsep}{1.5pt}
\begin{tabular}{c|c|c c c c c}
 \hline   
 Model & Method &  Top-1 Accuracy $\uparrow$ & Mistake Severity $\downarrow$ & HierDist@1 $\downarrow$ & HierDist@5 $\downarrow$ & Hierdist@20 $\downarrow$ \\
 \hline
 \multirow{4}{*}{Type-I} & \cellcolor{blue!15}cross-entropy   &\cellcolor{blue!15}73.63 +/- 0.1165  &\cellcolor{blue!15}6.94 +/- 0.0208  &\cellcolor{blue!15}1.83 +/- 0.0117  &\cellcolor{blue!15} 5.70 +/- 0.0192  &\cellcolor{blue!15}7.34 +/- 0.0291 \\
 
 & \cellcolor{blue!15}CRM \cite{BetterMistakes_CRM}  &\cellcolor{blue!15}73.54 +/- 0.1495  &\cellcolor{blue!15}6.89 +/- 0.0272  &\cellcolor{blue!15}1.82 +/- 0.0155  &\cellcolor{blue!35}4.82 +/- 0.0062  &\cellcolor{blue!35}6.03 +/- 0.0041 \\
 
 & \cellcolor{blue!15}Flamingo \cite{flamingo}       &\cellcolor{blue!15}72.34 +/- 0.1488  &\cellcolor{blue!15}6.93 +/- 0.0391  &\cellcolor{blue!15}1.92 +/- 0.0135  &\cellcolor{blue!15}5.75 +/- 0.0130  &\cellcolor{blue!15}7.41 +/- 0.0098 \\
 
 & \cellcolor{blue!15}HAFeature \cite{HAFeature}     &\cellcolor{blue!15}73.52 +/- 0.1613  &\cellcolor{blue!15}6.89 +/- 0.0281  &\cellcolor{blue!15}1.82 +/- 0.0125  &\cellcolor{blue!15}5.52 +/- 0.0176  &\cellcolor{blue!15}6.95 +/- 0.0120 \\
 
 \hline
 \multirow{5}{*}{Type-II} & cross-entropy  & 72.51 +/- 0.4317  & 6.95 +/- 0.0298  & 1.91 +/- 0.0338  & 5.69 +/- 0.0085  & 7.28 +/- 0.0082 \\
 
 & \cellcolor{blue!15}CRM \cite{BetterMistakes_CRM} &\cellcolor{blue!15}72.45 +/- 0.4077  &\cellcolor{blue!15}6.90 +/- 0.0274 &\cellcolor{blue!15}1.90 +/- 0.0308  &\cellcolor{blue!15}4.85 +/- 0.0090 &\cellcolor{blue!15}6.05 +/- 0.0057 \\
 
 & Flamingo \cite{flamingo} & 66.46 +/- 1.1572  & 7.05 +/- 0.0319  & 2.36 +/- 0.0921  & 5.77 +/- 0.0220  & 7.31 +/- 0.0120 \\
 
 & HAFeature \cite{HAFeature} & 68.32 +/- 0.9225  & 7.04 +/- 0.0356  & 2.23 +/- 0.0741  & 5.62 +/- 0.0223  & 6.97 +/- 0.0105 \\
 
 & \cellcolor{blue!15}HAFrame (ours) &\cellcolor{blue!35}74.00 +/- 0.3549 & \cellcolor{blue!35}6.89 +/- 0.0251 & \cellcolor{blue!35}1.79 +/- 0.0216 &\cellcolor{blue!15}4.94 +/- 0.0118 & \cellcolor{blue!15}6.15 +/- 0.0065 \\
\hline
\end{tabular}
\caption{Experiment results on tieredImageNet-H dataset. The details of type-I and type-II models are included in the training config.}
\label{tab:tiered_imagenet}
\end{table*}

\subsection{Ablation Study}
In this section, we examine the effectiveness of the proposed (1) transformation layer $\mathcal{T}(.)$; (2) the fixed HAFrame classifiers (dubbed as $HAF$); and (3) the cosine similarity-based auxiliary loss $\mathcal{L}_{COS}$ in a cumulative fashion with the type-I model as the baseline. All variants examined in this section are trained with the same settings introduced in the previous section, including the configuration of $\alpha$ and $\gamma$ for models with fixed HAFrame classifiers.  The results are shown in Table~\ref{tab:ablation2}. Our study shows that adding the transformation layer alone does not necessarily improve the model performance. However, fixing the corresponding classifier weights to a HAFrame improves the average mistake severity of three datasets. This also improves HierDist@5 and HierDist@20 for all four datasets. On top of these two changes, adding the auxiliary loss further facilitates the penultimate features to collapse on the HAFrame and reaches the best performance on all metrics among the variants examined except for a top-1 accuracy drop of 0.25\% on iNaturalist2019.

\begin{table*}[t]
\centering
\footnotesize\addtolength{\tabcolsep}{-3pt}
\begin{tabular}{c|c|c c c c|c c c c c}
\hline
 dataset & model & $\mathcal{L}_{CE}$ & $\mathcal{T}(.)$ & $HAF$ & $\mathcal{L}_{COS}$& Top-1 Acc $\uparrow$ & Mistake Severity $\downarrow$ & HieDist@1 $\downarrow$ & HieDist@5 $\downarrow$ & Hiedist@20 $\downarrow$ \\
 \hline 
\multirow{4}{*}{\shortstack{FGVC-\\Aircraft}} & ResNet50  &\cellcolor{green!15} \cmark &\cellcolor{red!15} \xmark &\cellcolor{red!15} \xmark &\cellcolor{red!15} \xmark & 79.18 +/- 0.5511  & 2.12 +/- 0.0240  & 0.44 +/- 0.0097  & 2.10 +/- 0.0033  & 2.67 +/- 0.0040 \\

& ResNet50* &\cellcolor{green!15} \cmark &\cellcolor{green!15} \cmark &\cellcolor{red!15} \xmark &\cellcolor{red!15} \xmark & 79.58 +/- 0.2727 & 2.15 +/- 0.0159 & 0.44 +/- 0.0067 & 2.11 +/- 0.0055 & 2.67 +/- 0.0034 \\

& ResNet50* &\cellcolor{green!15} \cmark &\cellcolor{green!15} \cmark &\cellcolor{green!15} \cmark &\cellcolor{red!15} \xmark & 79.18 +/- 0.5347 & 2.08 +/- 0.0299 & 0.43 +/- 0.0145 & 1.90 +/- 0.0056 & 2.55 +/- 0.0101 \\

& ours &\cellcolor{green!15} \cmark &\cellcolor{green!15} \cmark &\cellcolor{green!15} \cmark &\cellcolor{green!15} \cmark & \textbf{80.49 +/- 0.4692} & \textbf{2.02 +/- 0.0381} & \textbf{0.39 +/- 0.0039} & \textbf{1.74 +/- 0.0027} & \textbf{2.45 +/- 0.0024} \\

\hline 
\multirow{4}{*}{\shortstack{CIFAR-\\100}} & WideResNet28  &\cellcolor{green!15} \cmark &\cellcolor{red!15} \xmark &\cellcolor{red!15} \xmark &\cellcolor{red!15} \xmark & 77.65 +/- 0.2635  & 2.34 +/- 0.0271  & 0.52 +/- 0.0102  & 2.25 +/- 0.0084 & 3.19 +/- 0.0045 \\

& WideResNet28* &\cellcolor{green!15} \cmark &\cellcolor{green!15} \cmark &\cellcolor{red!15} \xmark &\cellcolor{red!15} \xmark & 76.45 +/- 0.2207  & 2.43 +/- 0.0235  & 0.57 +/- 0.0106  & 2.35 +/- 0.0049  & 3.30 +/- 0.0030 \\

& WideResNet28* &\cellcolor{green!15} \cmark &\cellcolor{green!15} \cmark &\cellcolor{green!15} \cmark &\cellcolor{red!15} \xmark & 77.30 +/- 0.3798 & 2.37 +/- 0.0131 & 0.54 +/- 0.0111 & 1.59 +/- 0.0108 & 2.71 +/- 0.0147 \\

& ours &\cellcolor{green!15} \cmark &\cellcolor{green!15} \cmark &\cellcolor{green!15} \cmark &\cellcolor{green!15} \cmark & \textbf{77.71 +/- 0.2319}  & \textbf{2.21 +/- 0.0108}  & \textbf{0.49 +/- 0.0066}  & \textbf{1.11 +/- 0.0018}  & \textbf{2.18 +/- 0.0013}  \\

\hline 
\multirow{4}{*}{\shortstack{iNatura-\\list2019}} & ResNet50  &\cellcolor{green!15} \cmark &\cellcolor{red!15} \xmark &\cellcolor{red!15} \xmark &\cellcolor{red!15} \xmark & 70.68 +/- 0.2097  & 2.22 +/- 0.0103  & 0.65 +/- 0.0068  & 1.95 +/- 0.0043  & 3.37 +/- 0.0040 \\

& ResNet50* &\cellcolor{green!15} \cmark &\cellcolor{green!15} \cmark &\cellcolor{red!15} \xmark &\cellcolor{red!15} \xmark & 70.44 +/- 0.1576  & 2.26 +/- 0.0071  & 0.67 +/- 0.0036  & 1.97 +/- 0.0060  & 3.40 +/- 0.0070 \\

& ResNet50* &\cellcolor{green!15} \cmark &\cellcolor{green!15} \cmark &\cellcolor{green!15} \cmark &\cellcolor{red!15} \xmark & \textbf{71.14 +/- 0.2245} & 2.19 +/- 0.0132 & 0.63 +/- 0.0025 & 1.39 +/- 0.0021 & 2.20 +/- 0.0048 \\

& ours &\cellcolor{green!15} \cmark &\cellcolor{green!15} \cmark &\cellcolor{green!15} \cmark &\cellcolor{green!15} \cmark &  70.89 +/- 0.1213       & \textbf{2.04 +/- 0.0107}               & \textbf{0.59 +/- 0.0033}        & \textbf{1.14 +/- 0.0033}        & \textbf{1.73 +/- 0.0023} \\

\hline 
\multirow{4}{*}{\shortstack{tiered-\\ImageNet\\-H}} & ResNet50  &\cellcolor{green!15} \cmark &\cellcolor{red!15} \xmark &\cellcolor{red!15} \xmark &\cellcolor{red!15} \xmark & 73.63 +/- 0.1165  & 6.94 +/- 0.0208  & 1.83 +/- 0.0117  & 5.70 +/- 0.0192  & 7.34 +/- 0.0291 \\

& ResNet50* &\cellcolor{green!15} \cmark &\cellcolor{green!15} \cmark &\cellcolor{red!15} \xmark &\cellcolor{red!15} \xmark & 72.51 +/- 0.4317  & 6.95 +/- 0.0298  & 1.91 +/- 0.0338  & 5.69 +/- 0.0085  & 7.28 +/- 0.0082 \\

& ResNet50* &\cellcolor{green!15} \cmark &\cellcolor{green!15} \cmark &\cellcolor{green!15} \cmark &\cellcolor{red!15} \xmark & 73.54 +/- 0.2328 & 6.93 +/- 0.0274 & 1.83 +/- 0.0117 & 5.45 +/- 0.0026 & 6.82 +/- 0.0048 \\

& ours &\cellcolor{green!15} \cmark &\cellcolor{green!15} \cmark &\cellcolor{green!15} \cmark &\cellcolor{green!15} \cmark & \textbf{74.00 +/- 0.3549} & \textbf{6.89 +/- 0.0251} & \textbf{1.79 +/- 0.0216} & \textbf{4.94 +/- 0.0118} & \textbf{6.15 +/- 0.0065} \\

\hline
\end{tabular}
\caption{The ablation study results for FGVC-Aircraft (1st row), and CIFAR-100 (2nd row), iNaturalist2019 (3rd row), tieredImageNet-H (4th row). Each row in the table corresponds to the average results of 5 runs with a 95\% confidence interval. Both ResNet50 and WideResNet28 are customized type-I models. The ResNet50*, WideResNet28*, and ours are customized type-II models.}
\label{tab:ablation2}
\end{table*}

\begin{figure*}[t]
\centering
\setlength{\tabcolsep}{8pt}
\begin{tabular}{c c c c}
FGVC-Aircraft & CIFAR-100 & iNaturalist2019 & tieredImageNet-H\\
\includegraphics[height=1.2in]{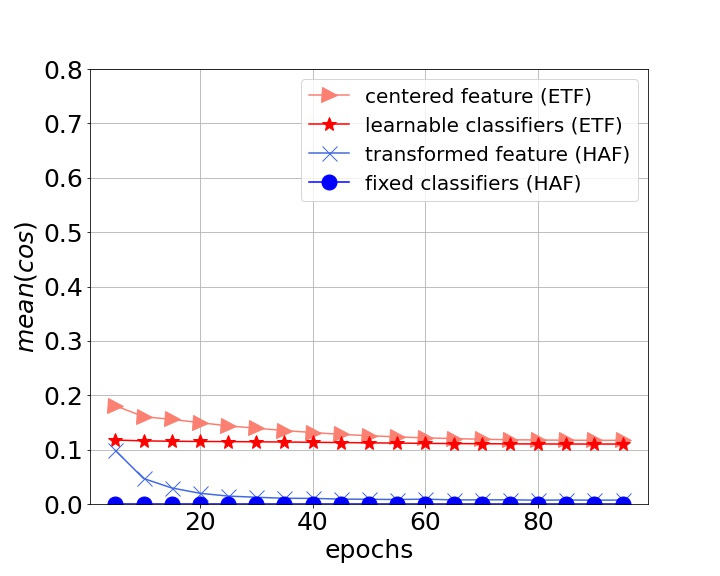} &
\includegraphics[height=1.2in]{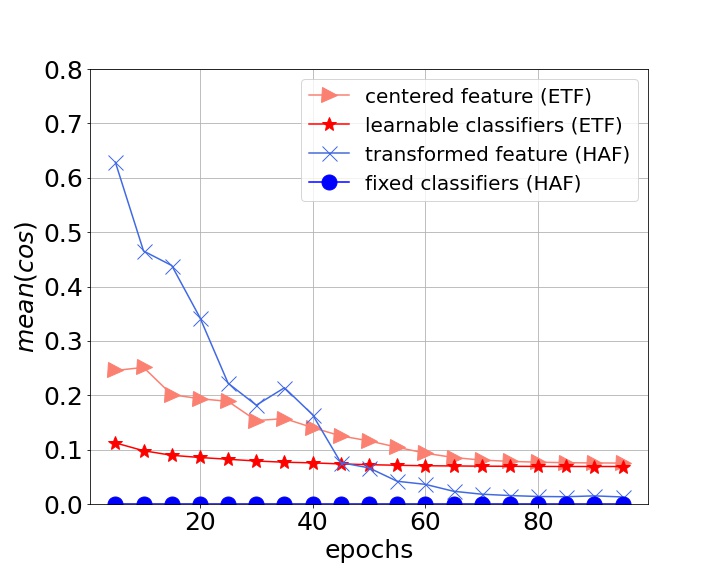} &
\includegraphics[height=1.2in]{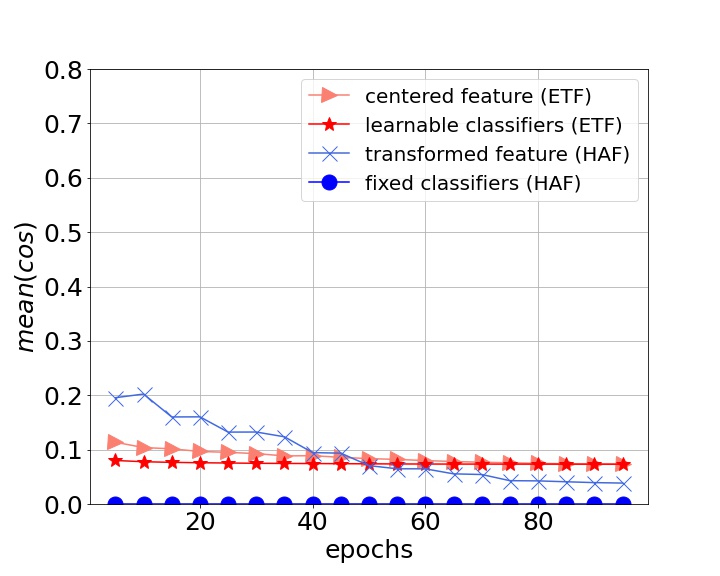} &
\includegraphics[height=1.2in]{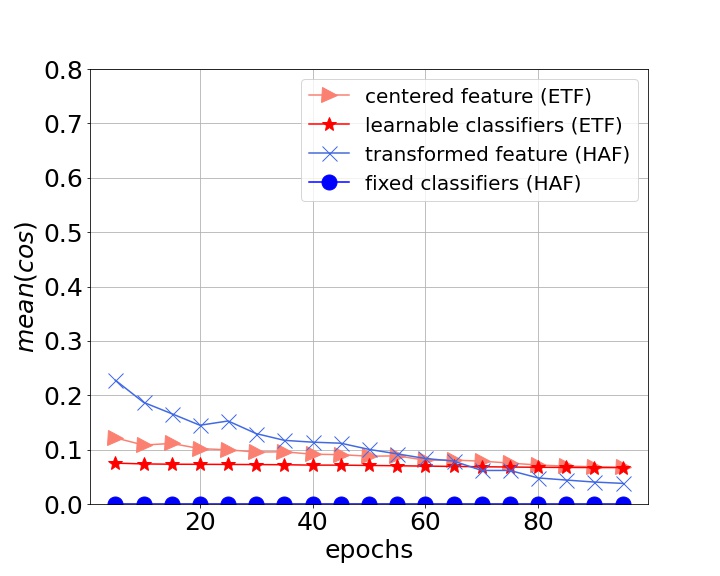} \\

\includegraphics[height=1.2in]{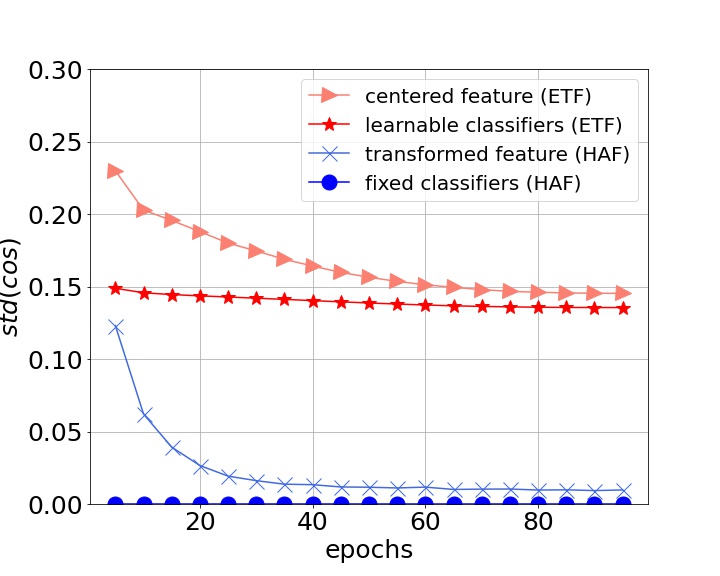} &
\includegraphics[height=1.2in]{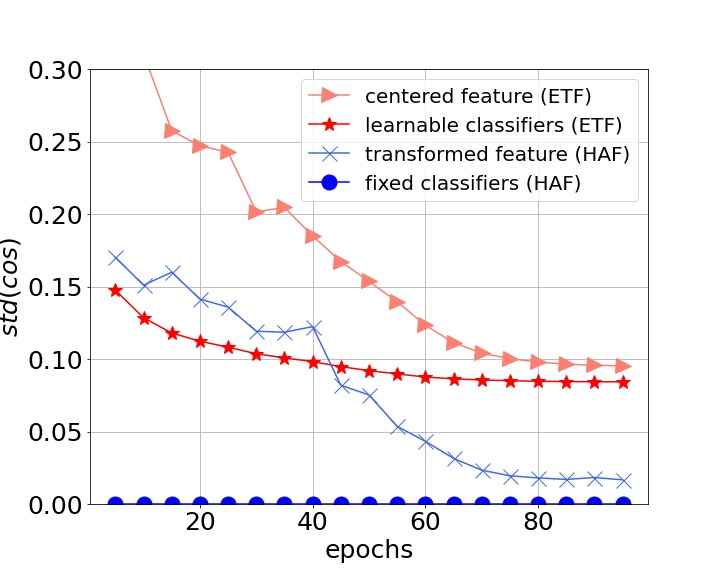} &
\includegraphics[height=1.2in]{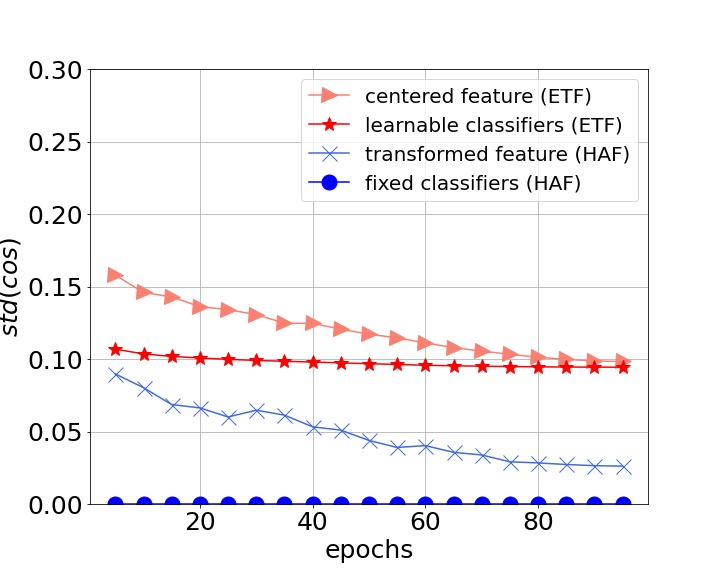} &
\includegraphics[height=1.2in]{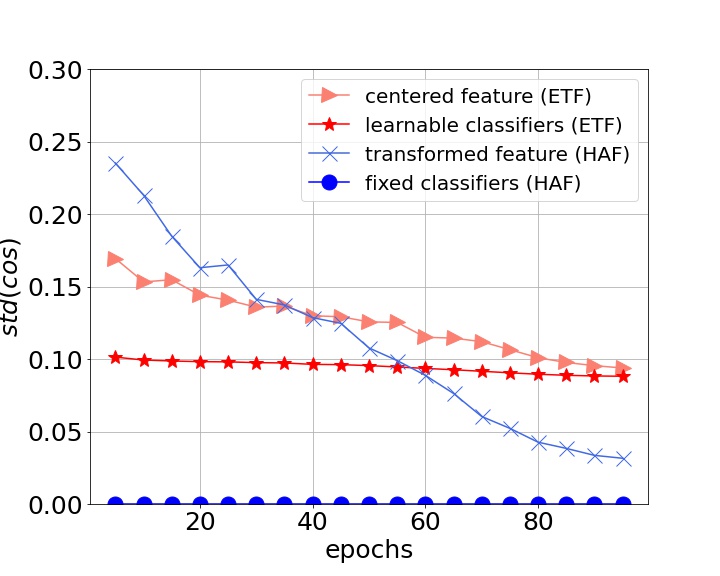} \\

\includegraphics[height=1.2in]{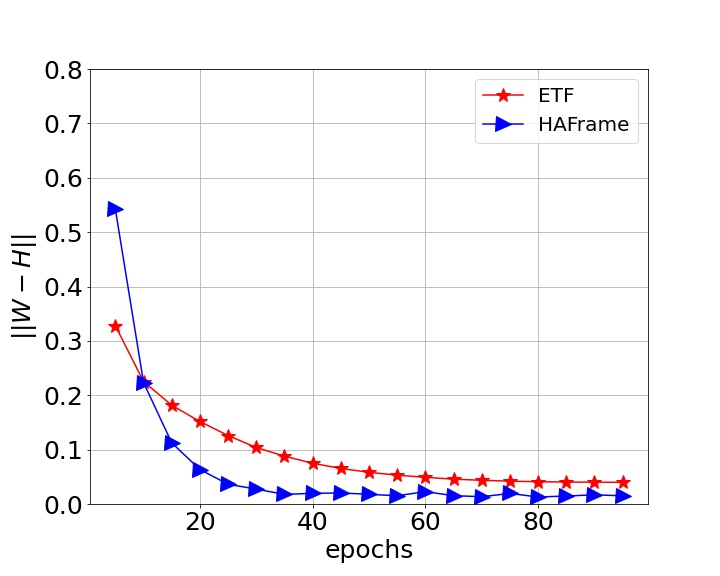} &
\includegraphics[height=1.2in]{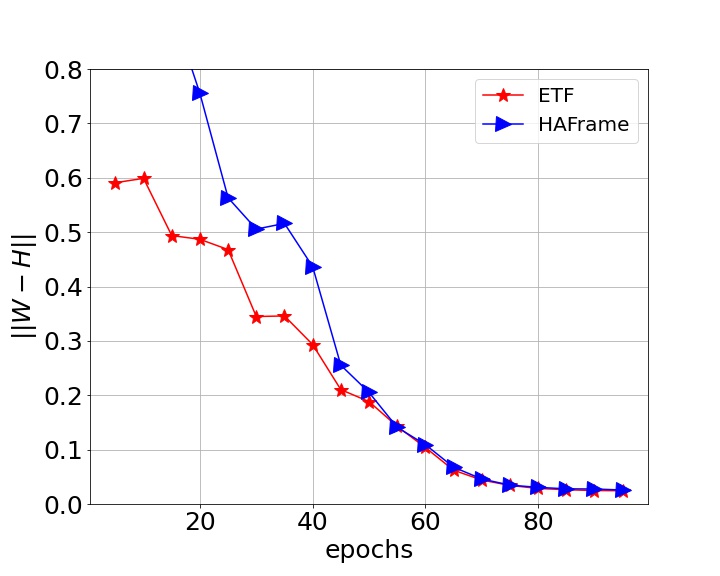} &
\includegraphics[height=1.2in]{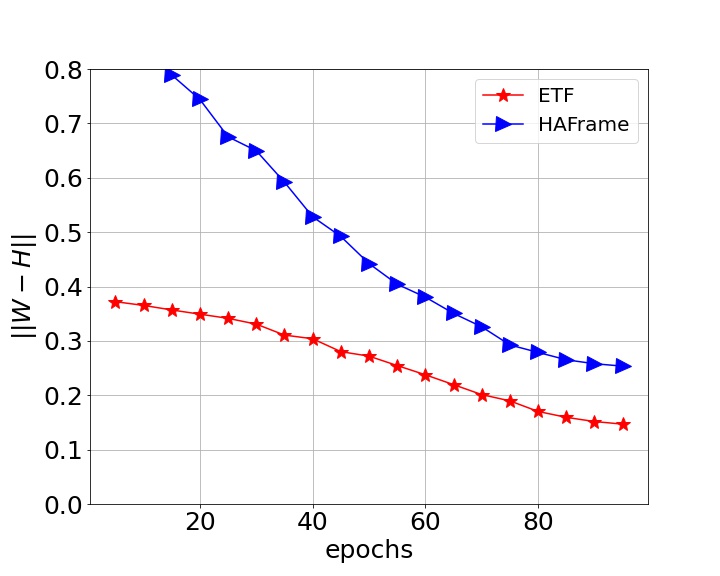} &
\includegraphics[height=1.2in]{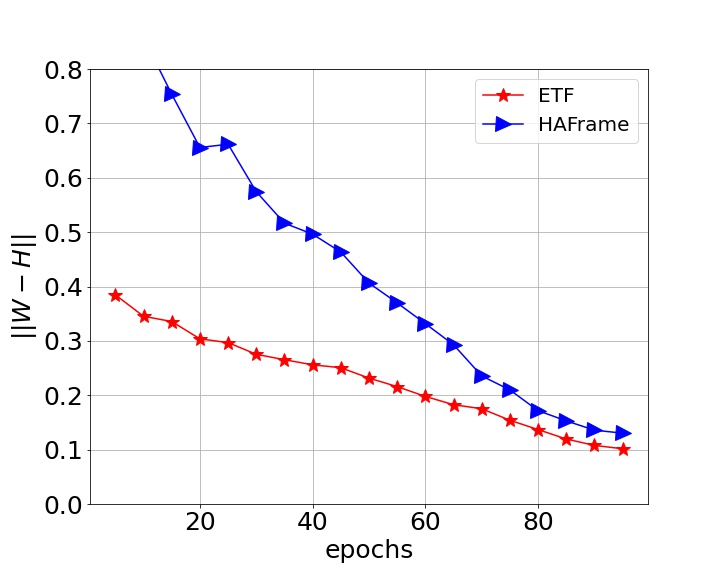}

\end{tabular}
\caption{Neural collapse visualization: HAFrame (blue) vs. ETF (red) on all four datasets. The x-axis is the epoch number, and the y-axis is either the mean (top row), the standard deviation (middle row), or the self-duality (bottom row) introduced in Sect.~\ref{sect: neural_collapse}. The number of hidden units in the transformation layer of type-I models is increased to 1010 and 608 for the respective models of iNaturalist2019 and tieredImageNet-H to meet the penultimate feature dimension's requirement ($d\geq K$) for ETF collapse.}
\label{fig: neural_collapse_viz}
\end{figure*}

\subsection{Neural Collapse on Hierarchy-Aware Frame} \label{sect: neural_collapse}

In this section, we briefly introduce the metrics employed to visualize the neural collapse on an ETF employed in \cite{neuralcollapse} and our extension to these metrics to visualize the collapse on a HAFrame. We compare our type-II models of fixed HAFrame classifiers with type-I baseline models of learnable classifiers from the perspective of HAFrame collapse and ETF collapse, respectively. We checkpoint the baseline and our models every five epochs during training of 100 epochs with the same settings as our previous experiments. 

\noindent 
\textbf{Angular collapse.} The classifiers and class means of training features should approach the ideal pair-wise angle during the neural collapse. Therefore, the respective pair-wise cosine similarities should approach the ideal cosine similarity $\hat{S}_{ij}$. We monitor the average of pair-wise cosine similarities during the training process to check if they are reaching the expected angle: 
\begin{equation} \label{eq: ETF_collapse_mean}
    Avg_{1\leq i<j\leq K}(|cos\angle(\boldsymbol{x}_i, \boldsymbol{x}_j) - \hat{S}_{ij}|)
\end{equation}
where $\boldsymbol{x}_i$, for $i=1,...,K$, are the set of $K$ vectors collapsing to the respective frame. They can either be the classifier weights or class means of the penultimate features (centered $\boldsymbol{\tilde{h}}$ for ETF, not centered $\boldsymbol{h}$ for HAFrame). We also track the standard deviation of pair-wise cosine similarities during the training process:
\begin{equation} \label{eq: ETF_collapse_std}
    Std_{1\leq i<j\leq K}(|cos\angle(\boldsymbol{x}_i, \boldsymbol{x}_j) - \hat{S}_{ij}|)
\end{equation}

For ETF collapse, the ideal cosine similarity between any pair of classes is equal to $\frac{-1}{K-1}$, i.e., $\hat{S}_{ij} = \frac{-1}{K-1}$ for $\forall i,j\in \{1,...,K\}$ and $i\neq j$ (equiangularity of ETF). 
The proposed HAFrame encodes the hierarchical relationship between classes into the pair-wise cosine similarities of the classes. Therefore, the ideal cosine similarity in Eqn.~\ref{eq: ETF_collapse_mean} and Eqn.~\ref{eq: ETF_collapse_std} is given by our mapping function in Eqn.~\ref{eq: mapping_hie_to_cos}, and we have $\hat{S}_{ij} = S_{ij}$ for HAFrame collapse. Both the mean and standard deviation in Eqn.~\ref{eq: ETF_collapse_mean} and Eqn.~\ref{eq: ETF_collapse_std} are approaching zero if neural collapse on an ETF or HAFrame occurs. 

The visualization results are shown in the top two rows of Fig.~\ref{fig: neural_collapse_viz}. Since we fixed our classifier to the pre-computed HAFrame, the associated means and standard deviations are always zeros.  
The features extracted from our HAFrame models reach smaller means (top row of Fig.~\ref{fig: neural_collapse_viz}) and standard deviations (middle row of Fig.~\ref{fig: neural_collapse_viz}) on all four datasets compared to the baseline, demonstrating the effectiveness of our approach.

\noindent
\textbf{Self-Duality.} During training, the class feature means of training examples should converge to the respective classifier vectors. Therefore, the means and classifiers become self-dual. Following \cite{neuralcollapse}, we visualize self-duality by measuring the Frobenius Norm of the difference between classifiers and class feature means:
\begin{equation}
    \left\Vert \frac{\boldsymbol{W}}{||\boldsymbol{W}||_F} - \frac{\boldsymbol{H}}{||\boldsymbol{H}||_F} \right\Vert_F
\end{equation}
where $\boldsymbol{W}$ is the classifier weight matrix, and $\boldsymbol{H}=[\boldsymbol{h}_1,...,\boldsymbol{h}_K]$ is a matrix of class feature means of training examples. This norm approaches zero as the features collapse to the associated classifiers. 

The visualization results are shown in the bottom row of Fig.~\ref{fig: neural_collapse_viz}. Both the baseline and our model reached self-duality on FGVC-Aircraft and CIFAR-100, and they are still approaching self-duality on the larger two datasets. We observed that more training epochs (e.g., 200 or 350 epochs) lead to better self-duality on the larger datasets with minor or no performance improvements.

\section{Conclusion}
Our proposed approach maps the pair-wise hierarchical distances of the flat classes into their associated cosine similarities and provides an analytical solution to the proposed hierarchy-aware frame for a given similarity matrix. The proposed approach is easy to implement as it only requires an extra 1x1 conv layer, a plug-in transformation layer, and freezing of the respective classifier to a HAFrame. Therefore, it is also easy to adapt to different hierarchies. Our approach offers a new route to reduce the mistake severity of model predictions from the neural collapse point of view. We examined the proposed approach through extensive experiments and our approach consistently reaches the best mistake severity while maintaining competitive classification accuracy (best on 3/4 datasets) and average hierarchical distances (best on 3/4 datasets). Future work may seek loss functions that better facilitate the collapse of penultimate features on the HAFrame or further optimize the architecture of the transformation layer to improve performance.

{\small
\bibliographystyle{ieee_fullname}
\bibliography{main}
}

\newpage

\begin{appendices}
\section{Prove the proposed hierarchy-aware frame satisfies the frame condition} \label{sect:appendix_a}
\noindent 
First, we repeat the definition of the frame condition \cite{frameCond} in Euclidean space:
\begin{definition}[Frame]
A set of vectors $\{\boldsymbol{\varphi}_k\}_{k=1}^M$ in $\mathbb{R}^N$ is called a frame for $\mathbb{R}^N$, if there exist constants $0< A \leq B < \infty$ such that
\begin{equation}\label{eq:frame_cond}
A||\boldsymbol{x}||^2\le \sum_{k=1}^{M}|\innerproduct{\boldsymbol{x}}{\boldsymbol{\varphi}_k}|^2\leq B||\boldsymbol{x}||^2,\:\forall \boldsymbol{x}\in \mathbb{R}^N
\end{equation}
where $\innerproduct{.}{.}$ and $||.||$ denote the dot product on $\mathbb{R}^N$ and its corresponding norm. 
\end{definition}

\noindent
Next, we prove the following theorem, i.e., our proposed HAFrame satisfies the above frame condition (Eqn.~\ref{eq:frame_cond}).

\begin{theorem}
The proposed HAFrame $\boldsymbol{W}$ (Eqn.~\ref{eq: HAF_solution}) satisfies the frame condition (Eqn.~\ref{eq:frame_cond}) if the associated cosine similarity matrix $\boldsymbol{S}$ in Eqn.~\ref{eqn: s=wTw} is positive definite.
\end{theorem}

\begin{proof}
To prove the proposed HAFrame $\boldsymbol{W}$ satisfies the frame condition \cite{frameCond} given in Eqn.~\ref{eq:frame_cond}, we substitute the set of vectors $\{\boldsymbol{\varphi}_k\}_{k=1}^{M}$ in the Eqn.~\ref{eq:frame_cond} with the set of vectors $\{\boldsymbol{w}_i\}_{i=1}^{K}$ consisting the HAFrame $\boldsymbol{W}$. It's equivalent to proving the following condition is satisfied where both $A$ and $B$ are positive constants: 
\begin{equation} \label{eq:prove_cond}
    A||\boldsymbol{x}||^2\leq \sum_{i=1}^{K}|\innerproduct{\boldsymbol{x}}{\boldsymbol{w}_i}|^2\leq B||\boldsymbol{x}||^2,\:\forall \boldsymbol{x}\in\mathbb{R}^K
\end{equation}
Therefore, we need to prove the following:
\begin{equation} \label{eq:matrix_vector}
    A\boldsymbol{x}^T\boldsymbol{x} \leq
    \sum_{i=1}^{K}(\boldsymbol{x}^T\boldsymbol{w}_i)(\boldsymbol{w}_i^T\boldsymbol{x}) \leq 
    B\boldsymbol{x}^T\boldsymbol{x}
\end{equation}
Since $\boldsymbol{W}=[\boldsymbol{w}_1\:\boldsymbol{w}_2\:...\:\boldsymbol{w}_K]$, we can further write $\sum_{i=1}^{K}(\boldsymbol{x}^T\boldsymbol{w}_i)(\boldsymbol{w}_i^T\boldsymbol{x})$ as:
\begin{align} 
    \sum_{i=1}^{K}(\boldsymbol{x}^T\boldsymbol{w}_i)(\boldsymbol{w}_i^T\boldsymbol{x}) = \boldsymbol{x}^T 
    \begin{bmatrix} 
    \boldsymbol{w}_1 & \boldsymbol{w}_2 & \cdots & \boldsymbol{w}_K 
    \end{bmatrix} 
    \begin{bmatrix}
        \boldsymbol{w}^T_1 \\
        \boldsymbol{w}^T_2 \\
        \vdots \\
        \boldsymbol{w}^T_K \\
      \end{bmatrix}
      \boldsymbol{x}
\end{align}
Therefore, we have:
\begin{equation} \label{eq:vector_eq}
    \sum_{i=1}^{K}(\boldsymbol{x}^T\boldsymbol{w}_i)(\boldsymbol{w}_i^T\boldsymbol{x}) = (\boldsymbol{x}^T\boldsymbol{W})(\boldsymbol{W}^T\boldsymbol{x}) = 
    \boldsymbol{x}^T\boldsymbol{W}\boldsymbol{W}^T\boldsymbol{x}
\end{equation}
Combining inequalities in Eqn.~\ref{eq:matrix_vector} and Eqn.~\ref{eq:vector_eq}, we have:
\begin{equation} 
    A\boldsymbol{x}^T\boldsymbol{x} \leq
    \boldsymbol{x}^T\boldsymbol{W}\boldsymbol{W}^T\boldsymbol{x} \leq 
    B\boldsymbol{x}^T\boldsymbol{x}
\end{equation}
If $||\boldsymbol{x}||=\boldsymbol{0}$, the \textit{frame condition} in Eqn.~\ref{eq:prove_cond} is met. Otherwise, we have:
\begin{equation}
    A\leq 
    \frac{\boldsymbol{x}^T\boldsymbol{W}\boldsymbol{W}^T\boldsymbol{x}}{\boldsymbol{x}^T\boldsymbol{x}}
    \leq B
\end{equation}
where $\frac{\boldsymbol{x}^T\boldsymbol{W}\boldsymbol{W}^T\boldsymbol{x}}{\boldsymbol{x}^T\boldsymbol{x}}$ is the Rayleigh quotient $R(\boldsymbol{W}\boldsymbol{W}^T,x)$  \cite{matrix_analysis_RayleighQuotient} for a real symmetric matrix $\boldsymbol{W}\boldsymbol{W}^T$, and it is bounded by the minimum and maximum eigenvalues of $\boldsymbol{W}\boldsymbol{W}^T$. Therefore, we have:
\begin{equation}
    \lambda_{min}\leq 
    \frac{\boldsymbol{x}^T\boldsymbol{W}\boldsymbol{W}^T\boldsymbol{x}}{\boldsymbol{x}^T\boldsymbol{x}}
    \leq \lambda_{max}
\end{equation}
where $\lambda_{min}$ and $\lambda_{max}$ are the smallest and largest eigenvalues of $\boldsymbol{W}\boldsymbol{W}^T$. Hence, we only need to prove that $\boldsymbol{W}\boldsymbol{W}^T$ is a positive definite matrix, i.e., for any $\boldsymbol{x}\neq \boldsymbol{0}$, we need to prove:
\begin{equation} \label{eq: equivalent_prove}
    \boldsymbol{x}^T\boldsymbol{W}\boldsymbol{W}^T\boldsymbol{x} =
    (\boldsymbol{W}^T\boldsymbol{x})^T(\boldsymbol{W}^T\boldsymbol{x}) =
    ||\boldsymbol{W}^T\boldsymbol{x}||^2 > 0
\end{equation}
It is equivalent to prove $\boldsymbol{W}^T = \boldsymbol{Q}\boldsymbol{D}^{\frac{1}{2}}\boldsymbol{U}^T$ is invertible (i.e., $\boldsymbol{W}^T\boldsymbol{x}\neq \boldsymbol{0}$ when $\boldsymbol{x} \neq \boldsymbol{0}$). We can find the inverse of $\boldsymbol{W}^T$: 
\begin{equation}
    (\boldsymbol{W}^T)^{-1} = \boldsymbol{U}\boldsymbol{D}^{-\frac{1}{2}}\boldsymbol{Q}^T
\end{equation}
where $\boldsymbol{U}$ is orthonormal, $\boldsymbol{Q}$ and $\boldsymbol{D}$ are acquired via eigendecomposition of real symmetric positive definite similarity matrix $\boldsymbol{S}=\boldsymbol{Q}\boldsymbol{D}\boldsymbol{Q}^T$, therefore $\boldsymbol{Q}$ is also orthonormal, and $\boldsymbol{D}^{-\frac{1}{2}}$ is a real matrix (eigenvalues of $\boldsymbol{S}$ are all positive real numbers). Since $\boldsymbol{W}^T$ is invertible, $\boldsymbol{W}^T\boldsymbol{x}\neq \boldsymbol{0}$ when $\boldsymbol{x}\neq \boldsymbol{0}$, this finish proving of Eqn.~\ref{eq: equivalent_prove}. Therefore, the matrix $\boldsymbol{W}\boldsymbol{W}^T$ is positive definite, and our HAFrame $\boldsymbol{W}$ satisfies the \textit{frame condition} in Eqn.~\ref{eq:frame_cond}. The proof is completed.
\end{proof}
\end{appendices}

\end{document}